\documentclass[journal]{IEEEtran}

\usepackage{amsmath,amssymb}
\usepackage{textcomp}
\usepackage{stfloats}
\usepackage{verbatim}
\usepackage{mathtools}
\usepackage{booktabs}
\usepackage{multirow}
\usepackage{array}
\usepackage{tabularx}
\usepackage{threeparttable}

\usepackage{graphicx}
\usepackage[table]{xcolor}
\usepackage{colortbl}
\usepackage{hhline}

\usepackage{xcolor}   % 颜色支持（soul依赖）
\usepackage{soul}     % 1. 导入高亮包
\sethlcolor{red!6}    % 2. 将高亮色设为淡红色

\usepackage[caption=false,font=normalsize,labelfont=sf,textfont=sf]{subfig}

\usepackage[compatibility=false]{caption}
\usepackage{algorithm}
\usepackage{algorithmic}
\usepackage{enumitem}
\usepackage{pifont}
\usepackage{hyperref}
\usepackage{cite}
\usepackage{url}

\AtBeginEnvironment{table}{\setlength{\tabcolsep}{3pt}}
\AtBeginEnvironment{table*}{\setlength{\tabcolsep}{3pt}}

\definecolor{headercolor}{HTML}{DBDBDB}
\definecolor{altrowA}{HTML}{EDEDED}
\definecolor{altrowB}{HTML}{FFFFFF}
\definecolor{lastrowcolor}{RGB}{226,239,213}
\definecolor{blue}{RGB}{237,237,237}
\definecolor{bluegray}{RGB}{255,255,255}
\definecolor{gree}{RGB}{226,239,213}
\definecolor{greegray}{RGB}{255,255,255}
\definecolor{purple}{RGB}{246,246,246}
\definecolor{purplegray}{RGB}{255,255,255}
\definecolor{textblue}{RGB}{0,0,0}

\begin{document}

\title{Mining Multi-Modality Spatio-Temporal Cues for Video Important Person Identification}

% NOTE: Replace the author block with real author information for journal submission.
\author{Xiao Wang, Minglei Yang, Bin Yang, Wenke Huang, Zheng Wang~\IEEEmembership{Senior Member,~IEEE}, Xin Xu~\IEEEmembership{Senior Member,~IEEE}, Mang Ye~\IEEEmembership{Senior Member,~IEEE}
% \thanks{This work was supported by National Natural Science Foundation of China under Grants (62501428,62302351,62225113)}
\thanks{X. Wang, M. Yang and X. Xu are School of Computer Science and Technology, Wuhan University of Science and Technology, Wuhan, Hubei 430065, China and Hubei Province Key Laboratory of Intelligent Information Processing and Real-time Industrial System, Wuhan University of Science and Technology,Wuhan 430065, China (e-mail:wangxiao2021@wust.edu.cn; yml0608@wust.edu.cn; xuxin@wust.edu.cn)}
\thanks{B. Yang, Z. Wang, and M. Ye are with School of Computer Science, National Engineering Research Center for Multimedia Software, Hubei Key Laboratory of Multimedia and Network Communication Engineering, Wuhan University, China (e-mail: yangbin\_cv@whu.edu.cn; wangzwhu@whu.edu.cn; yemang@whu.edu.cn) (Corresponding authors: Bin Yang; Mang Ye)}
\thanks{W. Huang is with College of Computing and Data Science, Nanyang Technological University (e-mail:wenkehuang@whu.edu.cn)}
}
\markboth{IEEE Transactions on Pattern Analysis and Machine Intelligence (Under Review)}%
{Anonymous Submission: Mining Multi-Modality Spatio-Temporal Cues for Video Important Person Identification}

\maketitle

\begin{abstract}
Identifying key individuals in video scenes is essential for applications such as automated video editing and intelligent surveillance. Current methods primarily focus on static images and immediate visual cues, overlooking the rich spatio-temporal information in videos. This leads to the phenomenon of Temporal Importance Shift (TIS), wherein individuals deemed significant in early frames may be demoted as the entire temporal context is considered.
To address this, we introduce the Video Important Person (VIP) identification task, aimed at automatically identifying the most influential individuals in videos while providing textual rationales. We present Temporal-VIP, a large-scale rationale-annotated dataset consisting of 9,249 video segments across 11 categories with aligned importance rationales. To mitigate TIS, we develop the VIP-Net framework, which includes a Social Cue Encoder (SCE) for extracting multi-modal spatio-temporal cues, a Temporal Importance Rectifier (TIR) for hierarchical cue fusion and cross-modal alignment, and VIP Inference for ranking individuals. Experimental results show that VIP-Net achieves 67.3\% accuracy, significantly outperforming state-of-the-art models (37.5\%–53.9\%) and yielding a mean rationale similarity of 0.63 to ground truth through feature-guided LLM refinement.
\textit{The dataset and code are available at \textcolor{magenta}{https://huggingface.co/datasets/yml2002/Temporal-VIP.}}

% \textit{\textcolor{magenta}{The dataset and code will be made publicly available.}}
\end{abstract}


\section{Introduction}

\begin{figure}[!t]
  \centering
  \includegraphics[width=0.47\textwidth]{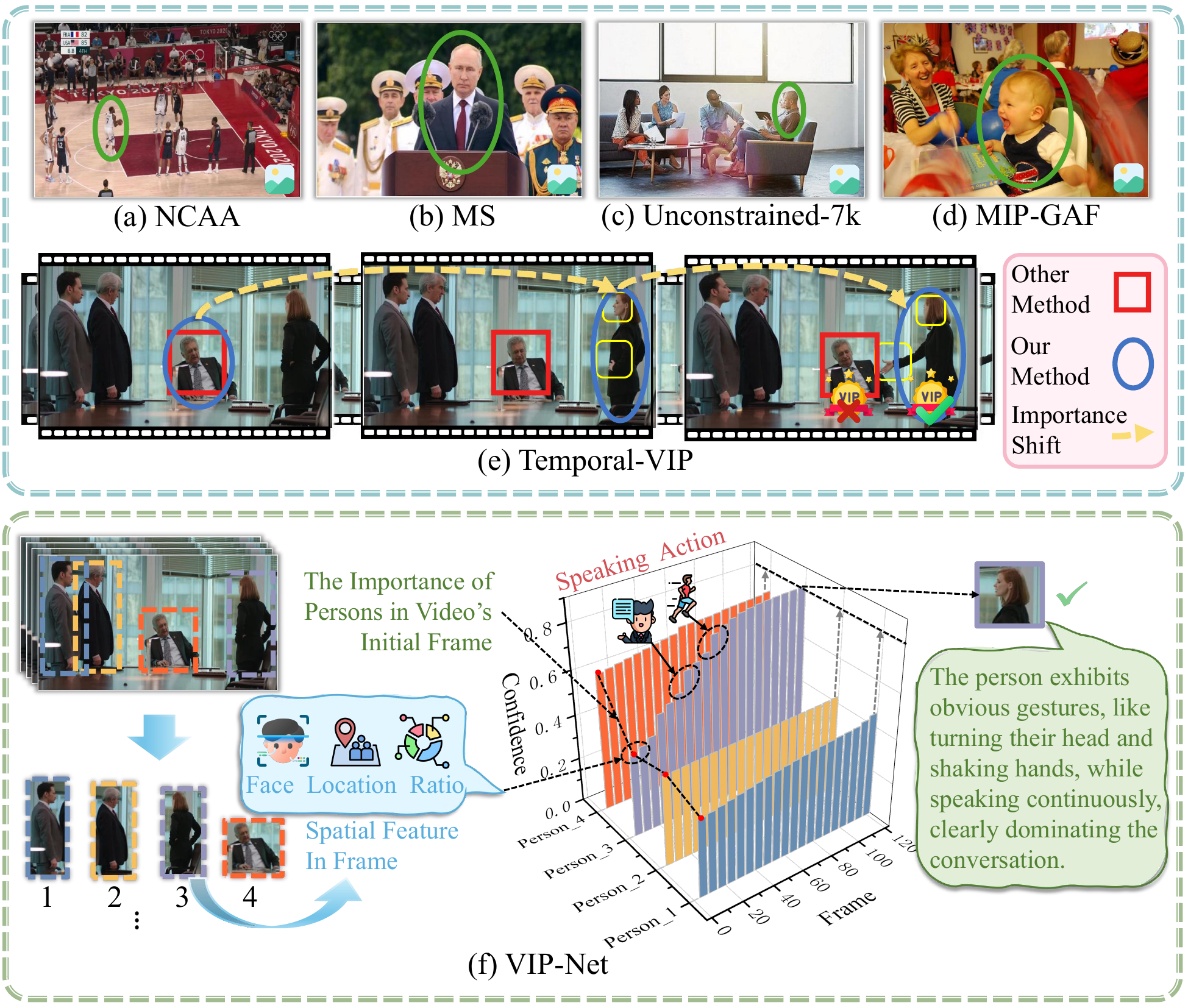}
  \caption{\textbf{Illustration of the Temporal Importance Shift (TIS) phenomenon and our motivation.} (a--d) Prior image-based important person (IP) datasets rely on instantaneous visual saliency (e.g., central positioning, large bounding box area). (e) In dynamic video sequences, individuals who are visually prominent in early frames (e.g., the seated person in the center) may be demoted as the temporal context unfolds. The true VIP (e.g., the standing speaker) emerges through sustained behavioral patterns such as speaking and gesturing. Red dashed boxes show erroneous detections by image-based baselines; green dashed boxes indicate our temporally-aware identifications. (f) Overview of our proposed VIP-Net framework.}
  \label{fig:motivation}
  \vspace{2mm}
\end{figure}

\IEEEPARstart{U}{nderstanding} human social dynamics is critical in computer vision, underpinning the development of Human-Centric and Artificial General Intelligence \cite{zhang2023temporal}. While current visual systems excel at recognizing atomic actions \cite{zhang2025bridgenet} and tracking individual trajectories \cite{zhu2025learning}, they struggle to elucidate implicit social hierarchies \cite{feng2024evolved}, particularly in identifying the most socially significant individual within multi-person scenarios \cite{yan2020higcin}. This process, termed Important Person Identification, is essential for applications such as intelligent surveillance, automated video editing \cite{sun2025ve,shang2025multitec}, sports broadcasting \cite{liu2022value}, and socially-aware human-robot interaction \cite{liu2025efficient}. These applications require intelligent agents to focus on social dynamics rather than mere visual prominence \cite{ye2021deep}.

To grasp the uniqueness of VIP (Video Important Person), it is vital to distinguish between traditional video understanding and social scene analysis. Person tracking \cite{zhang2024offsetnet} answers \textit{``Where is the person?''} and action recognition \cite{chi2024infogcn++} addresses \textit{``What are they doing?''} Conversely, VIP identification tackles the higher-order question: \textit{``Who drives the social event and why?''} For instance, in a corporate meeting, the VIP is often not the most centrally located individual but rather the one whose contributions significantly influence the discussion. Thus, accurately identifying key figures in real-world scenarios necessitates a nuanced understanding of social interactions, hierarchical dynamics, and contextual semantics. Historically, the community has predominantly approached this problem through the lens of static images. Foundational benchmarks such as the MS dataset~\cite{li2018personrank}, Unconstrained-7k~\cite{wang2021very}, and the recent MLLM-annotated MIP-GAF~\cite{madan2025mip} have driven significant progress. 

Consequently, prevailing methodologies predominantly rely on spatial cues and instantaneous visual saliency. Examples of such cues include central positioning, relative bounding box dimensions, and facial visibility. These features are commonly used to determine social importance, as demonstrated by existing image-based datasets shown in Fig.~\ref{fig:motivation} (a-d). However, from both sociological and cognitive perspectives, human social importance is inherently dynamic and evolves over time. It is expressed not only through static spatial prominence but also through sustained behavioral patterns, conversational dominance, and complex, evolving interactions. These factors go beyond the limitations of a single temporal snapshot. Therefore, depending solely on static frames forces models to adopt superficial spatial biases. This reliance fundamentally limits their ability to understand genuine social influence.

 This fundamental limitation becomes particularly evident when static paradigms are applied to real-world video sequences, revealing a critical bottleneck that we formally designate as the \textbf{Temporal Importance Shift (TIS)} problem. In complex social interactions, an individual's ``importance'' is seldom static; rather, it is a highly dynamic attribute that shifts non-linearly over time due to the evolution of the event, changes in interaction targets, and the transfer of semantic dominance. We conceptualize TIS as the phenomenon wherein the social focal point transitions among individuals as the temporal context develops. For instance, as illustrated in Fig.~\ref{fig:motivation} (e), within a meeting room scenario, a seated executive may initially appear visually prominent and function as the VIP. However, when a standing participant begins to articulate a critical point through speech and gesture, the semantic dominance of the event swiftly transfers to her, designating her as the true VIP for that specific temporal interval.

 The TIS phenomenon underscores that instantaneous visual saliency (e.g., central positioning within the frame) often diverges significantly from genuine long-term social significance. Image-based methods, which lack temporal memory and event-level reasoning capabilities, are particularly vulnerable to TIS. They tend to assign importance to the individual who is currently most visible, resulting in inconsistent, flickering, and ultimately erroneous identifications. Moreover, addressing the TIS problem is not merely an issue of temporal smoothing; it necessitates bridging a substantial \textit{semantic gap} between low-level pixel variations and high-level social concepts. Accurately tracking the trajectory of TIS requires a synergistic understanding of spatial positioning, temporal actions, and multi-agent interpersonal interactions across extended time horizons. 
  Despite the evident necessity for video-based analysis, the transition from static image important person (IP) identification to dynamic video analysis is hindered by two primary challenges:
  \begin{itemize}
  \item \hl{\emph{Challenge 1}: \textbf{Absence of dedicated video benchmarks with explanatory rationales.}} Existing image-based datasets (e.g., MS~\cite{li2018personrank}, Unconstrained-7k~\cite{wang2021very}) lack temporal sequences and merely provide spatial coordinates without comprehensive descriptive rationales. Conversely, while the video understanding community has developed large-scale datasets for action recognition (e.g., Kinetics~\cite{kay2017kinetics}), spatio-temporal action localization (e.g., AVA~\cite{gu2018ava}), and multi-object tracking (e.g., SportsMOT~\cite{cui2023sportsmot}), these benchmarks focus on ``what actions are occurring'' or ``where the people are.'' They lack semantic annotations~\cite{ma2023towards,zhai2024background} to annotate the hierarchical social structure or answer the higher-order question of ``who is socially important and why.''
  \item \hl{\emph{ Challenge 2}: \textbf{Architectural inadequacy in modeling long-term social relations.}} Our extensive empirical investigations reveal that directly adapting state-of-the-art multi-object trackers (e.g., ByteTrack~\cite{zhang2022bytetrack}, Wu et al.~\cite{wu2021track}), video saliency and feature-based detectors (e.g., Samba~\cite{he2025samba}, MGFN~\cite{chen2023mgfn}, Han et al.~\cite{han2024mutuality}), or even powerful Multimodal Large Language Models (e.g., BLIP-2~\cite{li2023blip}, TinyLLaVA~\cite{zhou2024tinyllava}) yields sub-optimal performance. These models either lack explicit inter-personal relationship modeling or fail to align long-term behavioral cues with spatial prominence, making them highly susceptible to TIS. Furthermore, they operate as black boxes, unable to provide transparent justifications for their social predictions.
  \end{itemize}

  To address \textbf{Challenge~1}, we propose the novel task of \textbf{Video Important Person (VIP) identification}. Supporting this task, we introduce the \textbf{Temporal-VIP} benchmark. 
  It consists of 9,249 meticulously curated video clips spanning 11 diverse real-world social categories, including speeches, meetings, and interviews. These clips capture rich and dynamic multi-person interactions. Beyond standard annotations, Temporal-VIP provides a comprehensive suite of labels. It supplies frame-level important person annotations alongside per-individual tracking identifiers and bounding boxes. Moreover, the benchmark includes textual rationales that explicitly explain the social basis for each importance judgment, thereby facilitating research in explainable visual understanding.

In the domain of social understanding, merely outputting a localization bounding box constitutes an insufficient solution; fostering genuine trust necessitates that artificial intelligence systems articulate the underlying reasoning behind their decisions. This imperative propels the field beyond pure perception towards the paradigms of Explainable Artificial Intelligence and multimodal reasoning. Consequently, a significant gap is revealed between prevailing video understanding methodologies and the specific requirements of visually important person identification. Existing benchmark approaches, such as JRDB-Act~\cite{ehsanpour2022jrdb} and MSR-VTT~\cite{xu2016msr}, predominantly focus on modeling collective group activities or generating general video descriptions, rather than capturing the nuanced dynamics of individual importance evolution over time.

  To tackle \textbf{Challenge~2} and fundamentally resolve the TIS problem, we propose Video Important Person identification Network (\textbf{VIP-Net}), a principled framework designed to explicitly reason over spatio-temporal social cues and track the dynamic transfer of semantic dominance. The core intuition behind VIP-Net is inspired by human cognitive processes: we first perceive atomic behavioral cues and then contextualize them within a global relational layout to rectify early visual biases caused by TIS. Specifically, VIP-Net features a Social Cue Encoder (SCE) that extracts complementary spatial (e.g., centrality, area) and temporal (e.g., action, lip movement) evidence for each individual. To prevent the model from being misled by instantaneous saliency shifts, we introduce a novel Temporal Importance Rectifier (TIR). The TIR performs hierarchical multi-modal fusion and inter-personal relational modeling across the entire video sequence, effectively capturing the trajectory of importance transfer and ensuring robust VIP identification even in the presence of severe TIS. 
  
  A natural question arises: \textit{Why not simply rely on the emergent capabilities of modern Multimodal Large Language Models (MLLMs) to resolve TIS?} Our preliminary explorations revealed that while MLLMs excel at describing objective visual elements and general actions within a video, they lack the deep-level cognitive reasoning required to comprehend complex social hierarchies and determine who the true important person should be. When faced with long videos exhibiting severe TIS, pure MLLMs often fail to grasp the underlying semantic dominance, hallucinating social relationships or missing subtle behavioral handovers. Therefore, we adopt a hybrid, feature-guided paradigm. VIP-Net acts as a specialized, deterministic engine that rigorously quantifies spatio-temporal social cues and models the TIS trajectory. Subsequently, a feature-guided LLM refinement module serves as a cognitive interpreter, translating these grounded representations into natural language rationales. This design ensures that our explanations are not only linguistically fluent but strictly anchored in visual evidence, providing transparent decision-making that bridges the gap between visual perception and linguistic explanation.
  
  The main contributions in this paper are threefold:

\begin{enumerate}[label=\large\ding{\numexpr181+\arabic*\relax}, leftmargin=*, labelsep=0.5em]
\item \textbf{Task and Data Contribution:} We formally extend important person identification from static images to the dynamic video domain. We construct and release \textbf{Temporal-VIP}, the first large-scale video benchmark (9,249 clips, 11 categories) equipped with multi-modal annotations and textual rationales, serving as a new testbed for social scene understanding.
\item \textbf{Methodology Contribution:} We propose \textbf{VIP-Net}, a novel framework that explicitly addresses the Temporal Importance Shift (TIS) through hierarchical spatial-temporal alignment and inter-personal relational modeling. It effectively integrates instantaneous visual saliency with long-term behavioral dynamics.
\item \textbf{Empirical Contribution:} We demonstrate significant performance gains, with VIP-Net achieving 67.3\% accuracy compared to 37.5\%--53.9\% for state-of-the-art models, while our feature-guided LLM refinement produces rationales with 0.63 mean similarity to ground truth.
\end{enumerate}

\section{Related Work}
\label{sec:related_work}

  \subsection{Person Importance Identification}
  The pursuit of identifying important persons originated in the static image domain, driven by the necessity to decode social hierarchies and visual attention in group photographs. Foundational benchmarks, such as the MS and NCAA Basketball datasets~\cite{li2018personrank}, catalyzed early research by formulating person importance as a relational-based ranking problem relying on spatial and appearance features, alongside early affect-aware formulations~\cite{ghosh2018role}. The field subsequently evolved with the introduction of the Unconstrained-7k dataset~\cite{wang2021very} and relation-centric refinements~\cite{Li2019LearningTL}, which expanded the scope to diverse, in-the-wild environments. To better capture complex interactions, methodologies advanced from basic feature aggregation to sophisticated relational modeling. For instance, JSRII-GNN~\cite{hong2020learning} employed graph neural networks to explicitly model social connections by treating individuals as nodes and their interactions as edges. Similarly, Wang et al.~\cite{wang2022towards} introduced structural causal models to disentangle genuine social importance from spurious visual correlations, while PI-Net~\cite{zhai2024background} explored the interplay between foreground subjects and background contexts. Most recently, the MIP-GAF dataset~\cite{madan2025mip} and IMPACT~\cite{rampuria2026impact} leveraged multimodal modeling~\cite{chang2024survey} and interpretability-oriented designs for more explainable important person analysis.
  
  Despite these significant strides, existing paradigms are fundamentally bottlenecked by their reliance on static images. They operate under the restrictive assumption that social importance can be reliably inferred from a single instantaneous snapshot \cite{he2017mask} (e.g., central positioning, relative bounding box size, or a frozen pose). When these image-based methods are naively adapted to dynamic video scenarios via frame-by-frame processing, they suffer from severe temporal inconsistencies. They fail to capture evolving social dynamics, such as sustained speaking, active gesturing~\cite{zhao2021space}, or the sudden handover of attention, inevitably falling victim to the Temporal Importance Shift (TIS) phenomenon. Our work addresses this critical gap by formally extending the task to the video domain and proposing a framework that explicitly models long-term behavioral cues.

  \subsection{Video Understanding and Social Dynamics}
  Current video analysis methodologies excel at fundamental perceptual tasks but fall short of high-level social reasoning. Instance-level perception models, including multi-object trackers (e.g., ByteTrack~\cite{zhang2022bytetrack}, SORT~\cite{wojke2017simple}, TrackFormer~\cite{meinhardt2022trackformer}, and recent MOT variants~\cite{cao2023observation,dendorfer2021motchallenge}) and action recognition frameworks (e.g., two-stream networks~\cite{feichtenhofer2019slowfast,carreira2017quo} and video transformers~\cite{arnab2021vivit,bertasius2021space,tong2022videomae}), can accurately localize individuals and classify atomic actions on standard benchmarks~\cite{kay2017kinetics,soomro2012ucf101,goyal2017something}. Recent human-centric video understanding models~\cite{chen2025motionllm,peng2025actionart} further strengthen behavior-level perception. Furthermore, recent advances in vision foundation models, such as DINOv2~\cite{oquab2023dinov2} and SAM~\cite{kirillov2023segment,liu2023referring}, provide robust, universal visual representations. However, these approaches primarily answer ``where the entities are'' or ``what actions occur,'' lacking the capacity to infer social hierarchies \cite{girdhar2019video}.
  
  Beyond basic perception, Spatio-Temporal Relational Modeling has garnered significant attention. Frameworks operating on datasets like Action Genome or AVA~\cite{gu2018ava} utilize Spatial-Temporal Graph Convolutional Networks (STGCN) to capture human-object and human-human interactions. Conversely, group-level perception tasks, such as collective activity recognition~\cite{bagautdinov2017social,wu2019learning,wang2022congnn} and social relation recognition~\cite{liu2019social,sun2017domain}, attempt to understand multi-agent dynamics. Yet, these architectures are inherently optimized for classifying predefined action categories or overall group behaviors, rather than evaluating the continuous, zero-sum transfer of social dominance. VIP identification serves as a crucial bridge between instance-level perception and group-level social understanding. Our proposed VIP-Net achieves this by leveraging robust spatio-temporal feature extractors and integrating them through a novel Temporal Importance Rectifier to explicitly model the dynamic transfer of semantic dominance.

  \subsection{Multimodal Large Language Models in Video}
  The advent of Multimodal Large Language Models (MLLMs) has revolutionized vision-language understanding~\cite{radford2021learning}. Models such as BLIP~\cite{li2022blip}, BLIP-2~\cite{li2023blip}, InstructBLIP~\cite{dai2023instructblip}, LLaVA~\cite{liu2024improved}, Qwen-VL~\cite{bai2023versatile}, TinyLLaVA~\cite{zhou2024tinyllava}, Chat-UniVi~\cite{jin2024chat}, and LLaVA-CoT~\cite{xu2025llava} have demonstrated remarkable zero-shot capabilities in image captioning and visual question answering. Building on earlier video-language pre-training pipelines~\cite{miech2019howto100m,zhu2020actbert,bain2021frozen,xu2021videoclip}, this paradigm has been extended to the video domain with models like VideoLLaMA 3~\cite{zhang2025videollama}, VideoChat-Flash~\cite{li2024videochat}, TimeChat~\cite{ren2024timechat}, Video-LLaVA~\cite{lin2024video}, LLaVA-ST~\cite{li2025llava}, and Video-ChatGPT~\cite{maaz2024video}, while recent large-scale foundation efforts~\cite{wang2024internvideo2} further improve multimodal video modeling. Related benchmark/tooling efforts and surveys~\cite{chen2024sharegpt4video,fu2025video,duan2024vlmevalkit,tang2025video}, together with socially-aware reasoning extensions~\cite{qin2026humansense}, also broaden the current MLLM landscape.
  
  While MLLMs possess strong semantic reasoning capabilities for general video description \cite{sun2019videobert}, our empirical evaluations reveal that they struggle with the fine-grained, multi-person relational reasoning required for VIP identification. Because they are predominantly trained on global video-text pairs, they lack the dense temporal grounding needed to track dynamic importance shifts among multiple interacting subjects. Furthermore, to handle the computational burden of long videos, most Video MLLMs employ aggressive token compression or temporal pooling strategies~\cite{qian2024streaming,ye2025re}. This inevitably discards fine-grained spatial details---such as a subtle shift in gaze or a brief hand gesture---which are often the critical tipping points for semantic dominance in social events. Consequently, pure MLLMs often suffer from ``attention drift,'' hallucinating social relationships or missing subtle behavioral handovers. In contrast, our approach utilizes a specialized, deterministic architecture (VIP-Net) for accurate spatio-temporal importance ranking, and employs MLLMs strictly in a feature-guided refinement capacity. This hybrid design ensures that the generated rationales are both linguistically natural and strictly anchored in extracted visual evidence.

  \subsection{Explainable and Rationale Generation}
  The necessity for artificial intelligence systems to provide transparent justifications has become increasingly critical, particularly in subjective tasks like social importance assessment. Traditional post-hoc explainability methods, such as Grad-CAM~\cite{selvaraju2017grad} and LIME~\cite{ribeiro2016should}, provide visual saliency maps for model decisions. While visual explainability provides a basic sanity check, it falls short in complex social scenarios where the ``why'' involves multi-step reasoning over time. Recent advances in understanding transformer reasoning~\cite{chefer2021transformer} and multimodal rationale generation~\cite{sun2024review}, including cross-modal approaches like LXMERT~\cite{tan2019lxmert}, have shown promise in generating textual justifications by aligning visual and textual representations.
  
In the domain of VIP identification, the mere output of a bounding box is insufficient for fostering user trust; it is imperative for the system to elucidate \textit{why} a particular individual is considered important. The generation of rationales for social tasks presents inherent complexities that exceed those associated with physical tasks \cite{yan2018spatial} (e.g., visual question answering regarding object colors), as it necessitates a multi-faceted justification that encompasses spatial cues, temporal patterns, and social dynamics \cite{dosovitskiy2020image}. This work advances the current understanding in this area by introducing a compositional rationale generation module that systematically translates deterministic spatio-temporal feature rankings into coherent natural language explanations. This approach effectively bridges the gap between low-level visual perception and high-level cognitive reasoning, thereby enhancing the interpretability and transparency of the VIP identification process.

\section{Methodology}
\label{sec:methodology}

\subsection{Task Definition}

We formulate video important person identification as a multi-modal ranking and text explanation task. Given a video containing multiple persons and textual descriptions, the objective is to: (1) identify the most socially significant individual, and (2) generate natural language descriptions for the importance identification.

\textbf{Problem Formulation.} Let $V \in \mathbb{R}^{T \times H \times W \times 3}$ denote a video sequence of $T$ frames with $N$ detected persons. Given bounding boxes $B \in \mathbb{R}^{T \times N \times 4}$, scene description $D_s$, and person descriptions $D_p = \{d_1, d_2, ..., d_N\}$, the task requires learning two functions.

\textbf{Importance Prediction.} We compute importance scores for each person and perform ranking to identify the most significant individual, which can be formulated as:
\begin{equation}
  \hat{y} = \mathop{\arg\max}_{n \in \mathcal{V}} \mathcal{P}\left( y=n \mid \mathbf{V}_{1:T}, \mathcal{B}, \mathcal{D}_{\text{spatial}}, \mathcal{D}_{\text{temporal}}; \mathbf{\Theta}_{\text{VIP}} \right),
\end{equation}
where $y \in \{1, 2, ..., N\}$ represents the identifier of the most important person, $\mathbf{V}_{1:T}$ denotes the video sequence, $\mathcal{B}$ denotes the bounding box trajectories, $\mathcal{D}$ represents the contextual descriptions, and $\mathbf{\Theta}_{\text{VIP}}$ embodies the trainable model parameters.

\textbf{Description Generation.} We generate natural language descriptions based on the predicted important person and their relative feature rankings within the scene, which can be defined as:
\begin{equation}
  \mathbf{E}^* = \mathop{\arg\max}_{\mathbf{E}} \prod_{l=1}^{L} P\left( e_l \mid e_{<l}, \mathbf{V}_{1:T}, \mathcal{R}(\hat{y}); \mathbf{\Phi}_{\text{LLM}} \right),
\end{equation}
where $E=\{e_1, \dots, e_L\}$ is an autoregressive rationale describing why person $\hat{y}$ is considered most important based on the percentile rank conditions $\mathcal{R}(\hat{y})$, and $\mathbf{\Phi}_{\text{LLM}}$ denotes the parameterized weights of the language model.

\textbf{Evaluation Protocol.}
The evaluation of importance prediction is conducted through ranking accuracy metrics, specifically Rank-1, Rank-2, and Rank-3 accuracy.

Description quality is assessed using SBERT~\cite{reimers2019sentence}-based cosine similarity between generated and ground truth descriptions, measuring semantic alignment of the description content.

% Importance prediction is evaluated using ranking accuracy metrics: Rank-1, Rank-2, and Rank-3 accuracy. Description quality is assessed using SBERT~\cite{reimers2019sentence}-based cosine similarity between generated and ground truth descriptions, measuring semantic alignment of the description content.

\begin{figure*}[!t]
    \centering
    \includegraphics[width=0.98\textwidth]{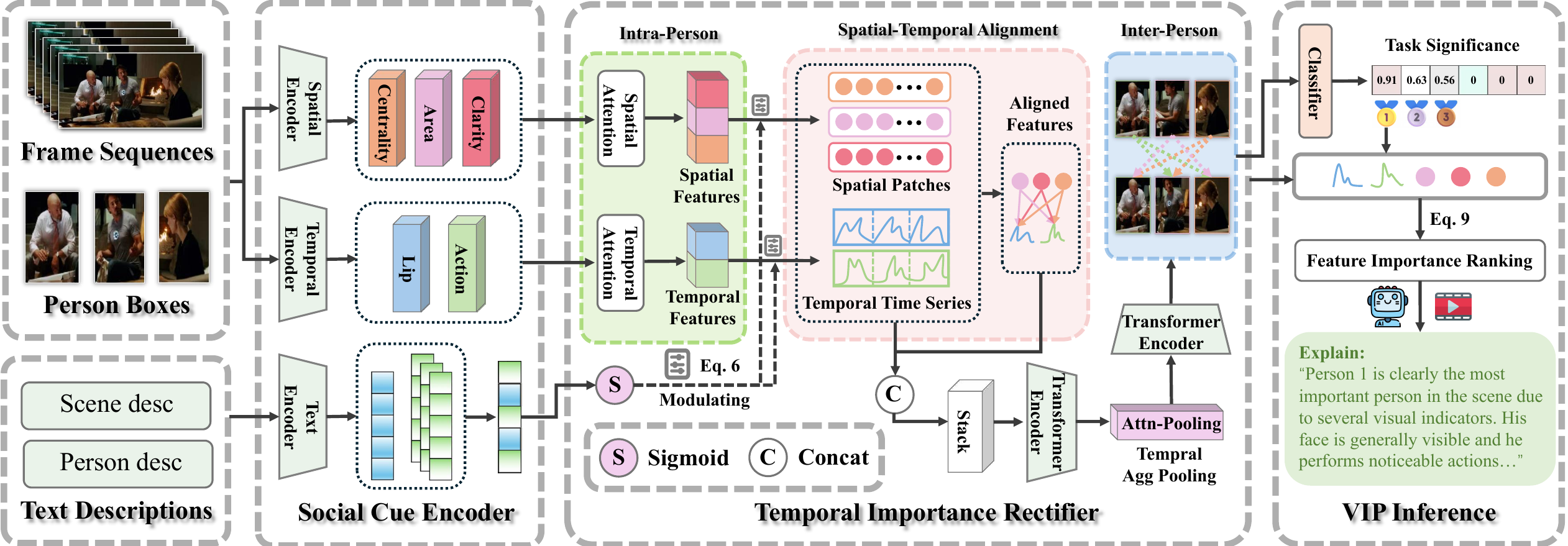}
  \caption{\textbf{Overall architecture of the proposed VIP-Net framework.} The framework consists of three core components: (1) \textit{Social Cue Encoder} extracts spatial visual features (centrality, area, clarity), temporal behavioral features (action, lip movement), and textual semantic features; (2) \textit{Temporal Importance Rectifier} addresses temporal importance shift through four hierarchical subsystems: intra-personal feature integration, contextual semantic modulation, spatial-temporal alignment, and inter-personal relations; (3) \textit{VIP Inference} generates importance rankings and produces natural language descriptions based on feature importance ranking.}
    \label{fig:overall_architecture}
\end{figure*}

\subsection{Social Cue Encoder}
We present a dual-dimensional encoding architecture that adeptly models social importance through the integration of complementary spatial and temporal evidence. This architecture systematically addresses the phenomenon of TIS by harnessing video-specific temporal dynamics that static methods cannot capture. Our approach elucidates the underlying mechanisms of person importance identification, where spatial positioning dictates attention allocation and temporal patterns provide insights into social roles. This comprehensive framework significantly advances the understanding of social importance within dynamic video contexts.

\textbf{Spatial Importance Encoding.} To effectively evaluate visual prominence, we design three interconnected spatial analyzers that operate in synergy. These analyzers collectively capture spatial cues that contribute to the identification of social importance, enabling a robust assessment of visual prominence within dynamic scenes.

\textbf{\textit{Centrality} }quantifies social focal points by assessing individuals' proximity to the visual center, positing a correlation between central positioning and compositional significance. This metric facilitates the identification of subjects possessing elevated social relevance, thereby reflecting their prominence within the visual hierarchy.
\begin{equation}
  \mathcal{S}_{\text{cen}}(n, t) = \max\left( 0, 1 - 2 \left\| \mathbf{p}^{(n,t)} - \mathbf{c}_{\text{img}} \right\|_2 \right),
\end{equation}
where $\mathbf{p}^{(n,t)}$ and $\mathbf{c}_{\text{img}} = (0.5, 0.5)^\top$ denote the normalized centroid coordinates and the image center. 

\textbf{\textit{Area}} assesses visual dominance by calculating the normalized bounding box area of subjects, effectively capturing their prominence in the foreground. This metric provides a quantitative measure of the visual weight that individuals carry within a scene, contributing to the overall evaluation of visual hierarchy.

\begin{equation}
  \mathcal{S}_{\text{area}}(n, t) = \frac{(x_2^{n,t} - x_1^{n,t})(y_2^{n,t} - y_1^{n,t})}{W \times H},
\end{equation}
where $(x_1^{n,t}, y_1^{n,t})$ and $(x_2^{n,t}, y_2^{n,t})$ are the top-left and bottom-right corner coordinates of the bounding box for person $n$ at frame $t$, and $W, H$ represent the spatial dimensions of the video frame.

\textbf{\textit{Clarity}} evaluates encoding quality by measuring focus through Laplacian variance in face regions detected by MediaPipe~\cite{lugaresi2019mediapipe}. This metric posits that sharper and better-lit subjects are more likely to attract preferential attention, thereby enhancing the overall visual effectiveness and perceptual clarity of the content.

\textbf{Temporal Importance Encoding.} In this research, we implement two specialized temporal analyzers that are adept at decoding dynamic social engagement patterns.

\textbf{\textit{Action}} focuses on modeling gestural and movement patterns through spatio-temporal feature extraction utilizing a 3D ResNet architecture~\cite{feichtenhofer2019slowfast}. This approach incorporates temporal attention mechanisms to enhance the recognition of relevant motion dynamics over time. To analytically capture the behavioral magnitude across a continuous observation window spanning $T$ frames, we formulate the action intensity metric as:
\begin{equation}
\mathcal{S}_{\text{act}}(n) = \frac{1}{|\mathcal{T}_{\text{window}}|} \sum_{t \in \mathcal{T}_{\text{window}}} \left\| \mathcal{F}_{\text{3D-ResNet}}\Big(\mathbf{V}_{t:t+\delta}^{(n)}\Big) \mathbf{W}_{\text{act}} \right\|_2,
\end{equation}
where $\mathcal{F}_{\text{3D-ResNet}}$ denotes the spatial-temporal convolutional mapping across a discrete block of $\delta$ frames for person $n$, projecting raw optical motion into a feature embedding, subsequently parameterized by the transformation matrix $\mathbf{W}_{\text{act}}$.

\textbf{\textit{Lip Movement}} assesses conversational dominance by employing MediaPipe~\cite{lugaresi2019mediapipe} for lip keypoint tracking, coupled with Transformer-based temporal modeling~\cite{vaswani2017attention}. This methodology allows for a nuanced quantification of speaking frequency and intensity, thereby providing insights into the dynamics of verbal interaction within social contexts. We track the inter-frame geometric disparities of the bounding control points over sequential timesteps, defining conversational salience as:
\begin{equation}
\mathcal{S}_{\text{lip}}(n) = \text{TRM}_{\text{temporal}} \left( \bigoplus_{t=1}^{T-1} \lambda_{t} \Big| \mathbf{p}_{\text{upper}}^{(n, t)} - \mathbf{p}_{\text{lower}}^{(n, t)} \Big| \right),
\end{equation}
where $\mathbf{p}_{\text{upper}}^{(n, t)}$ and $\mathbf{p}_{\text{lower}}^{(n, t)}$ represent the spatial coordinates of the upper and lower lip anchor boundaries at frame $t$, $\lambda_t$ represents the learned temporal modulation decay factor, and $\bigoplus$ signifies the sequential integration sequence fed into the Temporal Transformer.

\textbf{Contextual Semantic Encoding.} In this study, we utilize the Bidirectional Encoder Representations from Transformers (BERT) model~\cite{devlin2019bert} to process descriptive data pertaining to scenes and individuals. This approach facilitates the extraction of semantic context, thereby enhancing visual analysis by integrating situational understanding. Specifically, given a textual description sequence $D = \{w_1, w_2, \dots, w_L\}$, we formulate the semantic contextualization as a deep self-attention mapping:
\begin{equation}
    \mathbf{F}_{\text{text}} = \text{MLP}\left( \frac{1}{L} \sum_{l=1}^{L} \text{TRM}\Big( \mathbf{E}_{w_l} + \mathbf{E}_{pos_l} \Big) \right) \in \mathbb{R}^{D},
\end{equation}
where $\mathbf{E}_w$ and $\mathbf{E}_{pos}$ denote the token and positional embeddings respectively, and $\text{TRM}$ represents the multi-layer Transformer encoder capturing the multi-grained propositional semantics. By leveraging BERT's capabilities, we aim to deepen the interpretation of visual content through a more comprehensive understanding of the contextual elements that influence social dynamics and interactions.

\subsection{Temporal Importance Rectifier}

We tackle TIS by employing four hierarchical subsystems, namely: intra-personal feature integration, spatial-temporal alignment, and inter-personal relations.

\textbf{Intra-Personal Feature Integration.} We model individual coherence by integrating sub-features within each modality through an independent, non-normalized gating mechanism \cite{wang2018non}. Given the $k$-th sub-feature representations $\mathbf{X}_m^{(k)} \in \mathbb{R}^{T \times N \times D}$ for modality $m$, we first project them into a latent evaluation space to formulate the adaptive contribution weights $\alpha_{m,k}$:
\begin{equation}
    \mathbf{F}_{\text{intra}}^{(m)} = \sum_{k=1}^{K_m} \sigma\left( \frac{ \left( \mathbf{X}_m^{(k)} \mathbf{W}_{v} + \mathbf{b}_{v} \right) \mathbf{q}_m^\top}{\sqrt{d_k}} \right) \mathbf{X}_m^{(k)},
\end{equation}
where $\mathbf{W}_{v}$ and $\mathbf{b}_{v}$ are the projection weights and bias, $\mathbf{q}_m \in \mathbb{R}^D$ is the modality-specific learnable focal query, $d_k$ is the scaling factor to stabilize gradients, $\sigma$ denotes the sigmoid activation ensuring independent gating scales across dimensions, and $K_m$ is the number of sensory sub-streams ($K_s=3, K_t=2$).

\textbf{Spatial-Temporal Alignment.} Context-dependent importance is dynamically modulated by injecting semantic priors from the text embedding $\mathbf{F}_{\text{text}}$. We utilize a multi-layer perceptron (MLP) to generate a dense semantic-aware gate, which is explicitly broadcasted across the temporal axis via the Kronecker product ($\otimes \mathbf{1}_T$) to modulate the intra-personal features:
\begin{equation}
    \mathbf{\Gamma}_{\text{gate}} = \sigma\Big( \text{MLP}(\mathbf{F}_{\text{text}}) \otimes \mathbf{1}_T \Big), \quad \mathbf{\tilde{F}}^{(m)}_{\text{intra}} = \mathbf{F}_{\text{intra}}^{(m)} \odot \mathbf{\Gamma}_{\text{gate}},
\end{equation}
where $\odot$ is the Hadamard product. Subsequently, to holistically align the temporally-dense dynamic behavioral cues $^{(d)}$ with spatial prominence anchors $^{(s)}$, we rigorously formulate a structured Multi-Head Cross-Attention (MHA) module. With $h$ denoting the number of attention heads, the $i$-th head $\mathbf{H}_i$ precisely interrogates dynamic sequences guided by spatial queries:
\begin{equation}
\begin{aligned}
    \mathbf{F}_{\text{align}} &= \text{LayerNorm}\left( \mathbf{\tilde{F}}^{(s)}_{\text{intra}} + \text{Concat}_{i=1}^h \left( \mathbf{H}_i \right) \mathbf{W}_O \right), \\
    \mathbf{H}_i &= \text{Softmax}\left( \frac{ \mathbf{Q}^{(s)}_i {\mathbf{K}^{(d)}_i}^\top}{\sqrt{d_h}} + \mathbf{M}_{p} \right) \mathbf{V}^{(d)}_i,
\end{aligned}
\end{equation}
where $\mathbf{W}_O$ is the output projection matrix, and $\mathbf{M}_p \in \{0, -\infty\}^{T \times T}$ acts as an explicit validity masking matrix strictly penalizing attention scopes to filter out padded regions or severe occlusion tracklets.

To systematically distill noise-resilient, instance-level semantic representations from the temporal dimensions, we finalize the representation through an Energy-based Temporal Attention Pooling:
\begin{equation}
    \mathbf{F}_{\text{person}}^{(n)} = \sum_{t=1}^{T} \left( \frac{\exp\left( \big( \mathbf{F}_{\text{align}}^{(t)} \mathbf{W}_p + \mathbf{b}_p \big) \mathbf{q}_p^\top \right)}{\sum_{\tau=1}^{T} \exp\left( \big( \mathbf{F}_{\text{align}}^{(\tau)} \mathbf{W}_p + \mathbf{b}_p \big) \mathbf{q}_p^\top \right)} \right) \mathbf{F}_{\text{align}}^{(t)},
\end{equation}
where the focal energy term $e_t = \big( \mathbf{F}_{\text{align}}^{(t)} \mathbf{W}_p + \mathbf{b}_p \big) \mathbf{q}_p^\top$ quantifies the temporal prominence at frame $t$. This probabilistic temporal aggregation firmly guarantees that $\mathbf{F}_{\text{person}}^{(n)}$ encapsulates the cohesive, long-term social interaction trajectory instead of being monopolized by fleeting, instantaneous visual anomalies.

\textbf{Inter-Personal Relations.} To model social relationships and relative importance among individuals, the person-level features are processed through a Transformer encoder:
\begin{equation}
  \mathbf{H} = \text{TransformerEnc}\left(\left[ \mathbf{F}_{\text{person}}^{(1)}, \mathbf{F}_{\text{person}}^{(2)}, \dots, \mathbf{F}_{\text{person}}^{(N)} \right]\right) \in \mathbb{R}^{N \times D_p},
\end{equation}
where $\mathbf{F}_{\text{person}}^{(i)}$ represents the relationally-pooled features for individual $i$, and $\mathbf{H}$ contains the relationally-enhanced representations. This mechanism enables each person's importance identification to consider the social context and comparative relationships within the group.

\subsection{VIP Inference}

The VIP inference module processes the relationally-enhanced person representations from the Temporal Importance Rectifier to generate final importance rankings. The person features are L2-normalized and fed through a classification head that outputs logits for each person. After temperature scaling, the module outputs probability distributions over all persons, predicting the inferred VIP distribution as:
\begin{equation}
  \mathcal{P}(\hat{y}=n \mid \mathbf{H}) = \frac{\exp\left( \frac{\mathbf{W}_c \mathbf{h}_n}{\tau_c \|\mathbf{h}_n\|_2} \right)}{\sum_{m=1}^{N} \exp\left( \frac{\mathbf{W}_c \mathbf{h}_m}{\tau_c \|\mathbf{h}_m\|_2} \right)},
\end{equation}
where $\mathbf{h}_n \in \mathbf{H}$ is the enhanced representation of person $n$, $\mathbf{W}_c$ denotes the classification projection weights, and $\tau_c$ serves as the temperature scaling hyperparameter controlling the sharpness of the probability distribution. The individual maximizing this probability is selected as the predicted VIP. Based on the identified VIP, we generate natural language descriptions through scene-aware feature ranking.

\textbf{Feature Importance Ranking.} Rather than using absolute feature values, we compute scene-relative rankings to highlight distinctive characteristics. For each feature type $k \in \{\text{centrality}, \text{clarity}, \text{area}, \text{lip movement}, \text{action}\}$ and the predicted VIP $\hat{n}$, we calculate the percentile rank:
\begin{equation}
  \mathcal{R}_{k}(\hat{n}) = \frac{1}{|\mathcal{V}| - 1} \sum_{m \in \mathcal{V} \setminus \{\hat{n}\}} \mathbb{I}\left( \mathcal{S}_{k}^{(m)} < \mathcal{S}_{k}^{(\hat{n})} \right), \quad k \in \mathcal{K},
\end{equation}
where $\mathcal{S}_k^{(m)}$ is the feature score for person $m$ on dimension $k$, $\mathbb{I}(\cdot)$ is the indicator function evaluating boolean conditions, and $\mathcal{V}$ denotes the set of valid reference persons.

\textbf{Rationales Generation.} We retain only cues whose percentile ranks $\mathcal{R}_{k}(\hat{n})$ exceed a margin threshold $\tau_m = 0.7$, indicating the VIP reliably outperforms at least 70\% of local interactants in that specific physical or social dimension. Each recursively retained rationale is verbalized through deterministic syntactic templates, forming coherent multi-dimensional justification vectors (e.g., \textit{``The person holds central prominence and engages in continuous active speech''}).

\textbf{LLM-Guided Refinement.} To enhance naturalness while maintaining alignment with the feature-based evidence, we post-process these structured descriptions using Qwen3-VL-8B-Instruct. Given the predicted VIP and feature-driven description, we extract the video clip and query the LLM. The refined rationale text $Y = \{y_1, y_2, \dots, y_M\}$ is probabilistically decoded through an autoregressive generation bounds maximized by the feature-guided conditioning:
\begin{equation}
\begin{aligned}
    P\Big(Y \mid \mathbf{V}, \mathcal{R}_{k}(\hat{n}), \text{prompt}\Big) &= \\
    \prod_{m=1}^{M} P_{\theta_{\text{LLM}}}\Big(y_m &\mid y_{<m}, \mathbf{V}, \mathbf{E}_{\text{prompt}} \oplus \mathbf{E}_{\text{cues}} \Big),
\end{aligned}
\end{equation}
where $\mathbf{E}_{\text{prompt}}$ and $\mathbf{E}_{\text{cues}}$ represent the instruction and discrete syntactic cues templates. The guided variant anchors reasoning to concrete evidence, while the unguided variant serves as ablation. Both use identical visual inputs, ensuring performance differences derive from feature guidance. The LLM is prompted to rephrase and expand the rule-based sentences into more natural and diverse expressions while preserving core reasoning logic.

\subsection{Training Objective}
We employ a multi-component training objective:
\begin{equation}
\begin{aligned}
  \mathcal{L}_{\text{total}} &= \underbrace{-\frac{1}{|\mathcal{B}|} \sum_{i \in \mathcal{B}} \log \hat{y}_i^{(n^*)}}_{\mathcal{L}_{\text{cls}}} + \lambda_{\text{text}} \underbrace{\left( 1 - \frac{\mathbf{v}_{t}^\top \mathbf{v}_{\hat{t}}}{\|\mathbf{v}_{t}\|_2 \|\mathbf{v}_{\hat{t}}\|_2} \right)}_{\mathcal{L}_{\text{text}}} \\
  &\quad + \lambda_{\text{cont}} \underbrace{\mathbb{E} \left[ -\log \frac{\exp(\mathbf{h}_v \cdot \mathbf{h}_v^+ / \tau_c)}{\sum_{\mathbf{h}^-} \exp(\mathbf{h}_v \cdot \mathbf{h}^- / \tau_c)} \right]}_{\mathcal{L}_{\text{cont}}} + \lambda_{\text{reg}} \|\Theta\|_2^2,
\end{aligned}
\end{equation}
where $\hat{y}_i^{(n^*)}$ is the predicted probability for the ground-truth VIP, $\mathbf{v}_{t}$ and $\mathbf{v}_{\hat{t}}$ are the SBERT embeddings of the ground-truth and predicted descriptions. For the contrastive formulation, $\mathbf{h}_v$ and $\mathbf{h}_v^+$ denote the anchor and positive pair features, $\tau_c$ is the temperature, and $\Theta$ encompasses all regularized model weights.

The classification loss $\mathcal{L}_{\mathrm{cls}}$ employs cross-entropy with mask-aware softmax normalization. The text similarity loss $\mathcal{L}_{\mathrm{text}}$ calculates cosine similarity between SBERT~\cite{reimers2019sentence} embeddings of predicted and ground-truth descriptions. The contrastive loss $\mathcal{L}_{\mathrm{cont}}$ employs InfoNCE~\cite{oord2018representation} to distinguish VIP features from non-VIP features within each scene, encouraging VIPs to have more similar representations while being distinct from non-VIPs. Standard L2 regularization $\mathcal{L}_{\mathrm{reg}}$ prevents overfitting. Textual descriptions serve as optional semantic priors via the contextual semantic branch, enhancing spatial-temporal features with scene-aware rationales when available; when absent, the gating reverts to identity mapping, enabling the model to function solely on visual signals. This design allows for flexible operation with or without textual inputs during validation and testing.

\textbf{Discussion.} VIP-Net utilizes spatial cues from image-based methods to identify key individuals and incorporates temporal processing to capture dynamic behaviors such as actions and speech. By merging spatial and temporal analyzes, VIP-Net improves its capacity to recognize and articulate the significance of individuals across varying contexts in video sequences, effectively addressing the issue of temporal importance shift (TIS) in real-world scenarios.

\section{Temporal-VIP Dataset}
\label{sec:dataset}

To address the critical lack of video-based benchmarks for social importance analysis, we introduce \textbf{Temporal-VIP}, the first large-scale, rationale-annotated dataset dedicated to Video Important Person identification. Unlike existing image-based datasets (e.g., MS~\cite{li2018personrank}, Unconstrained-7k~\cite{wang2021very}) that rely on instantaneous visual saliency, Temporal-VIP is explicitly designed to capture the dynamic evolution of social importance over time.

\subsection{Data Collection and Pre-processing}
The raw video data was meticulously sourced from 232 diverse, high-resolution videos, encompassing a wide spectrum of sources including cinematic movies, TV shows, documentary footage, and unconstrained YouTube clips. This diverse sourcing ensures a comprehensive coverage of real-world social interactions, with video resolutions ranging from 720p to 4K and frame rates between 24 and 60 fps. To construct the initial candidate clips, we employed a rigorous automated pipeline. First, we utilized the YOLOv8l-pose model to detect all individuals and their keypoints within the frames. Subsequently, ByteTrack~\cite{zhang2022bytetrack} was applied to associate these frame-level detections into continuous multi-object trajectories, utilizing both spatial IoU and appearance embeddings to handle brief occlusions. This automated process generated 64,145 initial video clips. We deliberately constrained the clip duration to be between 3 and 10 seconds; this temporal window is empirically optimal as it is sufficiently long to capture a complete, meaningful social interaction (e.g., a conversational turn, a collaborative action) while avoiding the semantic ambiguity that arises when multiple disjoint events are concatenated in excessively long videos.

These candidate clips then underwent a stringent manual screening process based on five criteria: (1) \textit{Detection and Tracking Integrity}: No missing detections or ID switches for the main subjects; (2) \textit{Visual Clarity}: Sufficient resolution and lighting to discern facial expressions and body language; (3) \textit{Social Interaction}: Presence of meaningful multi-person dynamics (e.g., conversations, collaborative tasks) rather than mere co-occurrence; (4) \textit{Temporal Dynamics}: The clip must exhibit temporal evolution, avoiding static, surveillance-style footage where nothing changes; (5) \textit{Absence of Severe Occlusion}: Key subjects must remain largely visible throughout the clip. This rigorous filtering reduced the dataset to 9,249 high-quality, validated clips.

\subsection{Annotation Protocol and Quality Control}
Given the inherent subjectivity in defining a ``socially important person,'' establishing a rigorous, objective, and reproducible annotation protocol is paramount. To mitigate individual biases, our core strategy was to deconstruct the abstract concept of ``social importance'' into a hierarchy of observable, quantifiable spatio-temporal behavioral cues. We recruited a dedicated team of annotators, prioritizing individuals with backgrounds in sociology or psychology to ensure a nuanced understanding of human interactions. The annotation process was structured into three rigorous phases:

\textbf{1. Protocol Definition and Calibration.} We established a comprehensive guideline that explicitly prohibits the exclusive reliance on static spatial biases, including the tendency to default to the individual located in the central position. Instead, annotators were trained to evaluate the significance of elements based on a rigorous hierarchy of evidence:
\begin{itemize}
\item Dynamic Behaviors: Conversational Dominance includes sustained speaking, active gesturing, and the capacity to capture the visual attention or gaze of others. Additionally, Significant Actions refer to the execution of tasks that are pivotal to the progression of an event, illustrated by examples such as a teacher writing on a board or an athlete executing a decisive play.
\item Context and Spatial: Scene-Specific Norms refer to contextual roles that are determined by the environment, such as the role of a judge in a courtroom. Spatial Prominence is characterized by central positioning or a large bounding box area. These cues are employed primarily in situations where primary behavioral signals are ambiguous.
\end{itemize}
Prior to the formal task, annotators completed a pilot study on a subset of 500 videos to calibrate their judgments and align their understanding of the cue hierarchy.

\textbf{2. Independent Multi-dimensional Annotation.} During the formal phase, each of the 9,249 video clips was independently processed by five annotators using a custom-developed interface supporting frame-by-frame playback. For each clip, annotators were required to perform two tasks: (a) identify the VIP's bounding box trajectory across the temporal window, and (b) explicitly select the primary rationales (from the predefined cues) that justified their decision. This dual-task design ensures that the identification is grounded in concrete visual evidence rather than intuition.

\textbf{3. Conflict Resolution and Quality Assurance.} To quantitatively assess annotation reliability, we measured the Inter-Annotator Agreement (IAA) using Fleiss' Kappa~\cite{mchugh2012interrater}. We achieved a high Kappa score of \textbf{0.89} (indicating ``almost perfect agreement''), demonstrating that our hierarchical cue protocol effectively standardized the subjective task. In cases where disagreements occurred---typically in highly complex, multi-agent scenarios with rapid semantic dominance shifts---a consensus was reached through a mandatory discussion panel led by a senior supervisor.

\textbf{4. Rationale Text Generation and Verification.} In addition to categorical cues, the provision of natural language rationales represents a significant advancement of Temporal-VIP. To achieve scalability, we employed the state-of-the-art multi-modal large language model known as Qwen2.5-VL. This model was prompted with video clips, human-annotated VIP IDs, and selected rationale cues to generate initial textual descriptions. Importantly, these machine-generated texts underwent a rigorous multi-level verification process conducted by annotators. This process involved correcting the texts across three critical dimensions: semantic accuracy by ensuring an accurate reflection of the VIP's role, grammatical fluency to maintain linguistic coherence, and strict visual alignment to eliminate any hallucinated actions.
\begin{table}[!t]
  \centering
  \caption{Comparison of Temporal-VIP with existing person importance datasets. T: Temporal, TD: Textual Description, IR: Importance Rationales.}
  \footnotesize
  \setlength{\tabcolsep}{4pt}
  \begin{tabularx}{\linewidth}{l >{\centering\arraybackslash}X *{3}{>{\centering\arraybackslash}X} r}
      \toprule
      \rowcolor{headercolor}
      \textbf{Dataset} & \textbf{Type} & \textbf{T} & \textbf{TD} & \textbf{IR} & \textbf{Scale} \\
      \midrule
      \rowcolor{blue}
      MS~\cite{li2018personrank} & Image & \ding{55} & \ding{55} & \ding{55} & 2,340 \\
      \rowcolor{bluegray}
      NCAA Basketball~\cite{li2018personrank} & Image & \ding{55} & \ding{55} & \ding{55} & 9,736 \\
      \rowcolor{blue}
      EMS~\cite{li2018personrank} & Image & \ding{55} & \ding{55} & \ding{55} & 6,067 \\
      \rowcolor{bluegray}
      Unconstrained-7k~\cite{wang2021very} & Image & \ding{55} & \ding{55} & \ding{55} & 7,250 \\
      \rowcolor{blue}
      CUC~\cite{wang2022towards} & Image & \ding{55} & \ding{55} & \ding{55} & 9,300 \\
      \rowcolor{bluegray}
  MIP-GAF~\cite{madan2025mip} & Image & \ding{55} & \ding{55} & \ding{51} & 16,550 \\
      \midrule
      \rowcolor{lastrowcolor}
      \textbf{Temporal-VIP (Ours)} & \textbf{Video} & \ding{51} & \ding{51} & \ding{51} & \textbf{9,249} \\
      \bottomrule
  \end{tabularx}
  \label{tab:dataset_comparison}
\end{table}

\begin{figure*}[!t]
  \centering
  \includegraphics[width=0.98\textwidth]{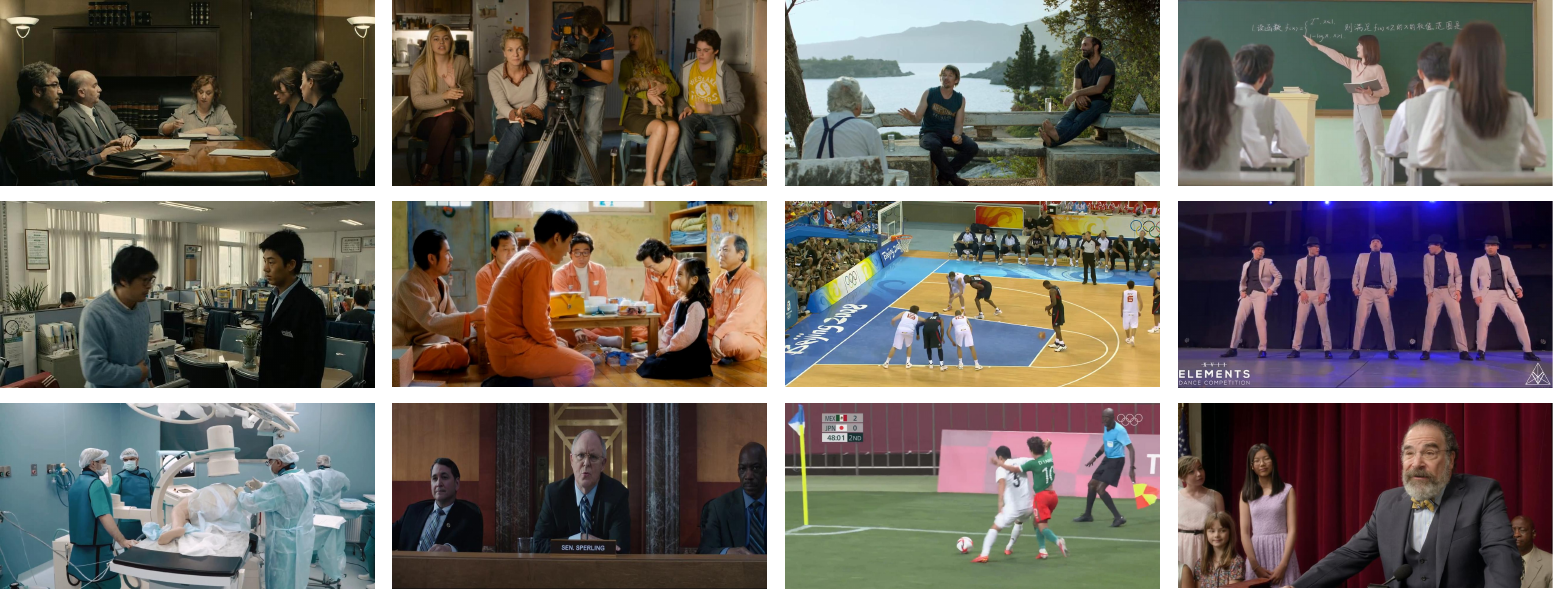}
  \caption{\textbf{Temporal-VIP dataset samples across 11 scene categories.} Representative video frames demonstrating diverse real-world scenarios including Office, Classroom, Conference, Restaurant, Sports, Interview, Performance, Public Space, Home, Courtroom, and Laboratory settings. Each sample showcases distinct VIP identification patterns where person importance assessment requires understanding both spatial positioning and temporal behavioral dynamics. The diversity of social contexts highlights the challenge of generalizing importance identification across varied scenarios.}
  \label{fig:dataset_samples}
\end{figure*}

\begin{figure}[!t]
  \centering
  \includegraphics[width=0.47\textwidth]{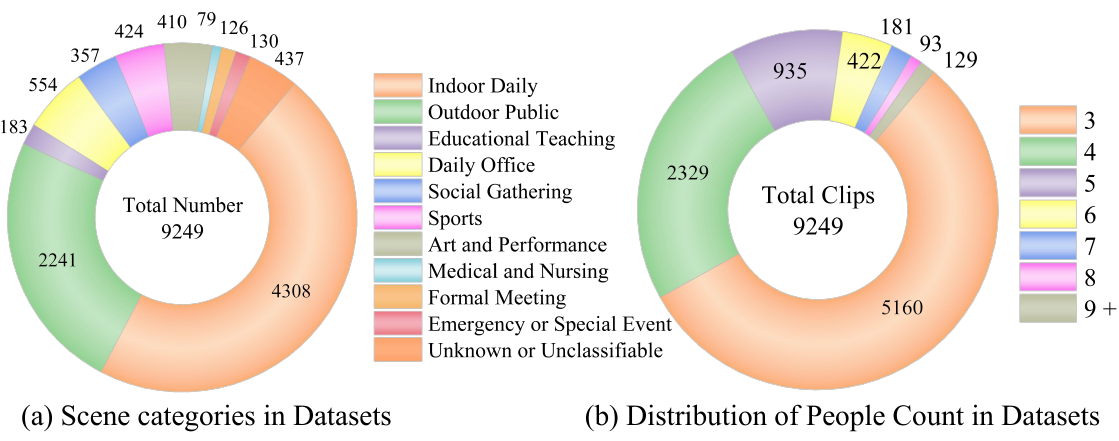}
  \caption{\textbf{Temporal-VIP dataset statistics.} (a) Distribution across 11 scene categories. (b) Distribution of number of people per video clip, showing that 3-person clips are the most frequent, followed by 4- and 5-person clips.}
  \label{fig:dataset_statistics}
\end{figure}

\subsection{Dataset Statistics and Diversity}
The finalized Temporal-VIP dataset comprises 9,249 video clips, systematically partitioned into 5,549 for training, 1,850 for validation, and 1,850 for testing. 
As illustrated in Table~\ref{tab:dataset_comparison}, Temporal-VIP is pioneering in its integration of the video modality with temporal analysis, extensive person descriptions, and clearly defined importance rationales. Fig.~\ref{fig:dataset_samples} showcases representative samples from the 11 distinct scene categories. These categories were carefully selected to cover a broad spectrum of social environments, ranging from highly structured settings with clear hierarchical norms (e.g., Courtroom, Conference, Classroom) to highly unconstrained, dynamic environments (e.g., Public Space, Sports, Party). This diversity ensures that models trained on Temporal-VIP must learn to generalize across different social rules rather than memorizing scene-specific biases.

Fig.~\ref{fig:dataset_statistics} details the distribution across these categories and the number of people per clip. The dataset predominantly features clips with 3 to 5 individuals. This distribution is intentional and aligns with sociological studies of small-group dynamics; groups of 3 to 5 people are large enough to exhibit complex social phenomena (such as shifting alliances, conversational turn-taking, and semantic dominance transfer) but small enough that individual behaviors remain distinct, avoiding the degradation into macroscopic crowd behavior analysis. Each clip is provided in a dual-format design: NPZ files containing standardized visual arrays and tracking metadata, and JSON files housing the hierarchical semantic annotations and rationales.

\section{Experiments}
\label{sec:experiments}

In this section, we comprehensively evaluate the proposed VIP-Net framework on the newly introduced Temporal-VIP dataset. We first detail the experimental setup, including implementation specifics and evaluation metrics. Subsequently, we benchmark VIP-Net against a diverse array of state-of-the-art methods, ranging from heuristic baselines to advanced multi-modal large language models, to demonstrate its superiority. We then present a fine-grained scene analysis to illustrate the model's robustness across various social contexts. Furthermore, we conduct extensive ablation studies to isolate and quantify the contribution of each architectural component, modality, and learning strategy. Finally, we evaluate the quality of the generated natural language rationales, validating the effectiveness of our feature-guided LLM refinement process.

\subsection{Experimental Setup}
\textbf{Implementation Details.} We implement the proposed VIP-Net framework using PyTorch and conduct all experiments on a workstation equipped with two NVIDIA RTX 4090 GPUs. The model is trained for 50 epochs with a batch size of 16. We employ the Adam optimizer with a base learning rate of $5 \times 10^{-5}$ and a weight decay of $1 \times 10^{-4}$. To ensure stable convergence, we utilize a cosine annealing learning rate schedule with a linear warmup over the first 5 epochs. For the multi-component loss function, the empirical weights are set to $\lambda_{\mathrm{text}} = 0.5$, $\lambda_{\mathrm{cont}} = 0.3$, and $\lambda_{\mathrm{reg}} = 0.0005$ based on our hyperparameter sensitivity analysis.

\textbf{Evaluation Metrics.} Following standard practices in important person identification~\cite{li2018personrank,wang2021very}, we evaluate the quantitative performance using Rank-1, Rank-2, and Rank-3 accuracy metrics. Rank-$k$ accuracy measures the percentage of test samples where the ground-truth VIP is present within the top-$k$ predicted candidates. To assess the quality of the generated natural language rationales, we employ the SBERT~\cite{reimers2019sentence} cosine similarity metric, which robustly measures the semantic alignment between the generated descriptions and the ground-truth rationales.

\subsection{Comparison with State-of-the-art Methods}

\subsubsection{Heuristic Baselines}
To establish the inherent complexity of identifying visually important persons and to substantiate the necessity of temporal reasoning, we initially evaluate three parameter‑free heuristic baselines. These baselines represent fundamental paradigms in visual importance assessment: the principle of \textbf{Maximum Centrality} based on geometric positioning, the principle of \textbf{Maximum Area} derived from visual attention, and the principle of \textbf{Maximum Face Clarity} guided by focus‑driven assessment.

Table~\ref{tab:heuristic_baselines} demonstrates that the Maximum Centrality baseline, which operates through geometric modeling, achieves the highest performance among all spatial methods with a Rank‑1 accuracy of 50.7\%, thereby surpassing purely attention‑based approaches. Nonetheless, VIP‑Net attains a substantial improvement of 16.6 percentage points over this geometric baseline. This result confirms that spatial positioning alone is insufficient to model the temporal dynamics that are essential for robust importance assessment in video.

\begin{table}[!t]
  \caption{Heuristic baseline performance on Temporal-VIP (Rank-1/2/3 accuracy, \%). Centrality achieves the best performance among spatial baselines through geometric modeling, while VIP-Net's temporal reasoning provides substantial improvements.}
  \centering
  \footnotesize
  \setlength{\tabcolsep}{3pt}
  \begin{tabularx}{\linewidth}{l *{6}{>{\centering\arraybackslash}X}}
    \toprule
    \rowcolor{headercolor}
    & \multicolumn{3}{c}{Overall} & \multicolumn{3}{c}{Indoor} \\
    \cmidrule(lr){2-4} \cmidrule(lr){5-7}
    \rowcolor{headercolor}
    \textbf{Method} & \textbf{R1} & \textbf{R2} & \textbf{R3} & \textbf{R1} & \textbf{R2} & \textbf{R3} \\
    \midrule
    \rowcolor{blue}
    Max Centrality & \underline{50.7} & 76.8 & 92.0 & \underline{49.6} & \underline{77.2} & 93.9 \\
    \rowcolor{bluegray}
    Max Area & 49.3 & \underline{76.9} & \underline{94.0} & 47.3 & 76.8 & \underline{94.8} \\
    \rowcolor{blue}
    Max Face Clarity & 32.7 & 61.8 & 84.4 & 33.1 & 64.9 & 86.1 \\
    \midrule
    \rowcolor{lastrowcolor}
   & \textbf{67.3} & \textbf{83.2} & \textbf{94.1} & \textbf{68.9} & \textbf{84.2} & \textbf{97.8} \\
    \rowcolor{lastrowcolor}
  \multirow{-2}{*}{\textbf{VIP-Net (Ours)}} & $\uparrow\,16.6$ & $\uparrow\,6.3$ & $\uparrow\,0.1$ & $\uparrow\,19.3$ & $\uparrow\,7.0$ & $\uparrow\,3.0$ \\
    \bottomrule
  \end{tabularx}
  \label{tab:heuristic_baselines}
\end{table}

\subsubsection{SOTA Method Adaptations \& Implementation Details}
Given the novel and emerging nature of the VIP identification task, there are no existing models directly tailored to solve it out-of-the-box. To establish a rigorous and comprehensive benchmark, we carefully adapt state-of-the-art methodologies from diverse paradigms. To ensure fairness, all learning-based baselines were re-trained or fine-tuned on the VIP dataset using their optimal hyper-parameter settings.

\noindent\textbf{1) Static Image IP Baseline:}
\begin{itemize}[leftmargin=*]
    \item \textbf{POINT}~\cite{li2018personrank}: As the seminal static image IP model, POINT infers importance based on spatial features and human-interactions within a single frame. We adapt it by extracting frame-level predictions across all video frames and utilizing a temporal majority-voting mechanism to aggregate the results into a video-level prediction. This serves to demonstrate the inherent limitations of neglecting temporal continuity in dynamic scenes.
\end{itemize}

\noindent\textbf{2) Video Feature Representation Baseline:}
\begin{itemize}[leftmargin=*]
    \item \textbf{MGFN}~\cite{chen2023mgfn}: Originally designed for video anomaly detection using grid-level features, we critically adapt MGFN for person-level ranking. Specifically, we extract person-level RoI (Region of Interest) features across the video sequence. The original frame-level anomaly loss is replaced with an individual-level Cross-Entropy loss, forcing its glance-and-focus mechanism to evaluate person importance based on long-term dependencies rather than anomaly magnitude.
    \item \textbf{Samba}~\cite{he2025samba}: A state-of-the-art Mamba-based framework for general video salient object detection (VSOD). We leverage its official predictive capabilities as a strong spatio-temporal saliency prior. To bridge the gap from pixel-level generic saliency to video person-level importance, we systematically adapt it via uniform keyframe sampling, frame-difference pseudo-flow generation for its original dual-stream input, heuristic intra-box saliency aggregation ($0.7 \times \text{mean} + 0.3 \times \text{max}$), and cross-frame top-$k$ temporal pooling.
\end{itemize}

\noindent\textbf{3) Heuristic Tracking Baseline:}
\begin{itemize}[leftmargin=*]
    \item \textbf{ByteTrack}~\cite{zhang2022bytetrack}: To evaluate whether low-level kinematics alone can solve the VIP identification task, we adapt this generic multi-object tracker. We define a heuristic scoring function that aggregates tracking metrics, ranking identities based on their spatial centrality, average bounding-box scale, and temporal trajectory persistence. This tests the strict hypothesis that importance correlates directly with continuous, prominent visibility.
\end{itemize}

\noindent\textbf{4) Vision-Language and Foundation Models:}
Large Vision-Language Models present strong zero-shot reasoning capabilities. We evaluated them by designing targeted prompts (e.g., \textit{``Analyze the social interaction in this scene. Which person (identified by bounding box [ID]) is the visual center or most important?''}) applied frame-by-frame, followed by temporal confidence aggregation.
\begin{itemize}[leftmargin=*]
    \item \textbf{X-CLIP}~\cite{ni2022expanding}: A video-text foundation model. We compute the contrastive similarity between the global video representation and text prompts constructed for each detected person.
    \item \textbf{BLIP-2}~\cite{li2023blip} \& \textbf{TinyLLaVA}~\cite{zhou2024tinyllava}: A heavier vision-language pre-training model (3.7B) and a highly efficient Multimodal Large Language Model (3B), respectively. Both are adapted via sequential visual prompting, explicitly asking the model to perform spatial reasoning and rank the visible identities based on interaction context.
\end{itemize}

\begin{table}[!t]
  \caption{Comparison with state-of-the-art methods across overall and indoor subsets. VIP-Net demonstrates significant gains over the top baseline (Rank-1/2/3 accuracy, \%). Bold formatting denotes the best results, while underlining indicates the runner-up. These findings underscore VIP-Net's superiority and the critical role of temporal reasoning in VIP identification.}
  \centering
  \footnotesize
  \setlength{\tabcolsep}{0pt}
    \begin{tabularx}{\linewidth}{l c *{6}{>{\centering\arraybackslash}X}}
        \toprule
        \rowcolor{headercolor}
        & & \multicolumn{3}{c}{Overall} & \multicolumn{3}{c}{Indoor} \\
        \cmidrule(lr){3-5} \cmidrule(lr){6-8}
        \rowcolor{headercolor}
        \textbf{Method} & \textbf{Reference} & \textbf{R1} & \textbf{R2} & \textbf{R3} & \textbf{R1} & \textbf{R2} & \textbf{R3} \\
        \midrule
        \rowcolor{blue}
        X-CLIP~\cite{ni2022expanding}        & ECCV'2022 & 23.0 & 50.1 & 80.5 & 24.4 & 53.0 & 84.5 \\
        \rowcolor{bluegray}
        BLIP-2~\cite{li2023blip}         & ICML'2023 & 37.5 & 48.7 & 56.2 & 36.6 & 47.6 & 54.9 \\
        \rowcolor{blue}
        TinyLLaVA~\cite{zhou2024tinyllava}     & arXiv'2024 & 49.5 & 70.9 & 89.9 & 46.1 & 71.0 & 90.5 \\
        \rowcolor{bluegray}
        ByteTrack~\cite{zhang2022bytetrack}     & ECCV'2022 & 49.7 & 64.6 & 74.5 & 47.8 & 62.1 & 71.6 \\
        \rowcolor{blue}
        MGFN~\cite{chen2023mgfn}          & AAAI'2023 & 50.5 & 67.1 & 87.4 & 50.7 & 67.9 & 90.2 \\
        \rowcolor{bluegray}
        Samba~\cite{he2025samba}      & CVPR'2025 & 52.0 & 77.8 & 92.1 & 51.2 & 78.9 & 94.0 \\
        \rowcolor{blue}
        POINT~\cite{li2018personrank} & FG'2018 & \underline{53.9} & \underline{79.1} & \underline{93.5} & \underline{51.8} & \underline{80.4} & \underline{95.1} \\
        \midrule
        \rowcolor{lastrowcolor}
          & & \textcolor{textblue}{\textbf{67.3}} & \textcolor{textblue}{\textbf{83.2}} & \textcolor{textblue}{\textbf{94.1}} & \textcolor{textblue}{\textbf{68.9}} & \textcolor{textblue}{\textbf{84.2}} & \textcolor{textblue}{\textbf{97.8}} \\
        \rowcolor{lastrowcolor}
        \multirow{-2}{*}{\textbf{VIP-Net (Ours)}} & \multirow{-2}{*}{\textemdash{}} & $\uparrow\,13.4$ & $\uparrow\,4.1$ & $\uparrow\,0.6$ & $\uparrow\,17.1$ & $\uparrow\,3.8$ & $\uparrow\,2.7$ \\
        \bottomrule
    \end{tabularx}
    \label{tab:main_results}
\end{table}

Table~\ref{tab:main_results} presents the comprehensive performance comparison. The results reveal several critical insights into the limitations of existing paradigms:
\begin{itemize}[leftmargin=*]
\item \textbf{The Failure of Video-Text Models:} X-CLIP performs poorly (23.0\% Rank-1). This indicates that global video-text alignment is insufficient for fine-grained, instance-level social reasoning.
\item \textbf{The Hallucination of MLLMs:} While TinyLLaVA (49.5\%) outperforms BLIP-2 (37.5\%), both MLLMs struggle. Our qualitative analysis shows they often hallucinate social relationships or overly rely on static spatial biases (e.g., always picking the person in the center), failing to track behavioral changes over time.
\item \textbf{The Limits of Tracking and Feature Detectors:} ByteTrack (49.7\%), MGFN (50.5\%), and Samba (52.0\%) achieve moderate performance. While Samba translates general video saliency effectively, their collective lack of explicit semantic behavioral modeling (like conversational turn-taking or specific semantic gestures) prevents them from accurately resolving complex social hierarchies.
\item \textbf{The Superiority of VIP-Net:} VIP-Net achieves a state-of-the-art 67.3\% Rank-1 accuracy, outperforming the best baseline (POINT, 53.9\%) by a massive 13.4\% absolute margin. This advantage is even more pronounced in Indoor scenes (68.9\% vs. 51.8\%, a 17.1\% improvement), where social interactions (e.g., meetings, classrooms) heavily rely on subtle temporal cues like speaking and gesturing, which VIP-Net explicitly models.
\end{itemize}
\subsection{Fine-grained Scene Analysis}
To further demonstrate the robustness of VIP-Net and understand the underlying mechanisms of social importance, we conduct a fine-grained analysis across the 11 distinct scene categories in our benchmark. We categorize these scenes into three primary social dynamics:

\textbf{1. Conversation-Driven Scenes (e.g., Meeting, Interview, Classroom):} In these highly structured environments, social importance is predominantly dictated by verbal communication and semantic dominance. Here, tracking-based methods (ByteTrack) and feature-based methods (MGFN, Samba) show severe degradation (often dropping below 45\% Rank-1 accuracy, e.g., Samba scores only 46.9\% on Daily and 38.6\% on Social scenes) because they rely on spatial movement or holistic magnitude, which is minimal in seated discussions. Conversely, VIP-Net maintains high accuracy (over 70\%) due to its dedicated lip movement module and temporal alignment, which effectively capture conversational turn-taking and sustained speaking.

\textbf{2. Action-Driven Scenes (e.g., Sports, Performance):} In these dynamic environments, physical action and spatial prominence dictate importance. While MLLMs like TinyLLaVA struggle due to rapid motion blur and complex multi-person occlusions, VIP-Net excels by leveraging its 3D ResNet action encoder. The model successfully identifies the key player making a decisive move or the lead performer, demonstrating that its temporal modeling extends beyond mere speech to encompass semantic physical actions.

\textbf{3. Unconstrained/Context-Driven Scenes (e.g., Public Space, Party):} These scenes exhibit the highest degree of Temporal Importance Shift (TIS), as social focal points change rapidly without strict rules. Baseline models frequently suffer from ``attention drift,'' greedily assigning importance to whoever temporarily moves to the center of the frame. VIP-Net's inter-personal relational modeling (via the Transformer encoder) proves crucial here, allowing it to contextualize brief spatial prominence against long-term behavioral patterns, thereby resisting spurious TIS.

\subsection{Qualitative Analysis of Temporal Importance Shift}
To explicitly validate our core claim---that VIP-Net essentially resolves the Temporal Importance Shift (TIS) problem---we qualitatively compare its temporal predictions against the strongest image-based IP baseline (POINT) and the most competitive video representation baselines (Samba and MGFN), specifically revisiting the motivating example shown earlier in Fig.~\ref{fig:motivation} (e).

As illustrated in this ``Conference'' scenario, during the initial frames of the video sequence, the seated executive (highlighted in the red box) is located centrally and occupies the spatial focal point. During this early static phase, baseline models correctly identify him as the VIP. However, as the temporal context unfolds, the standing woman on the right (highlighted in the blue oval) begins to speak and uses expressive hand gestures (indicated by the yellow boxes tracking her active movement). The true semantic dominance of the social event instantly transfers to her, making her the actual VIP of the clip. 

Faced with this TIS, the static baseline POINT continues to erroneously predict the seated executive, as its decision boundaries remain rigidly anchored to his spatial centrality. Similarly, the sequence-adapted MGFN and Samba baselines, while capable of aggregating temporal features and video-level saliency, fail to correctly re-assign the importance score because their generic feature magnitude evaluation cannot explicitly isolate and rank the sudden onset of conversational dynamics against spatial priors. In stark contrast, VIP-Net's Temporal Importance Rectifier successfully tracks this semantic handover. By integrating the woman's lip movement and action features with the contextual shift in the group's visual attention, VIP-Net decisively and stably updates its VIP trajectory to her. This perfectly aligns with multi-dimensional human cognitive assessment, proving our framework's capability to bridge the gap between static pixel saliency and dynamic social reasoning.

\subsection{Ablation Studies}

\begin{table}[!t]
  \caption{Ablation study of VIP-Net on Temporal-VIP, breaking down modality encoders, feature cues, fusion variants, and learning strategies (Rank-1/2/3 accuracy, \%). Each block isolates its component contribution, and bold numbers mark the best setting. RAFT denotes optical flow baseline. This ablation study confirms the critical role of temporal features, particularly lip movement, in achieving superior performance.}
  \centering
  \footnotesize
  \setlength{\tabcolsep}{1pt}
  \begin{tabularx}{\linewidth}{l l *{6}{>{\centering\arraybackslash}X}}
    \toprule
    \rowcolor{headercolor}
    & & \multicolumn{3}{c}{Overall} & \multicolumn{3}{c}{Indoor} \\
    \cmidrule(lr){3-5} \cmidrule(lr){6-8}
    \rowcolor{headercolor}
    \textbf{Category} & \textbf{Model Variant} & \textbf{R1} & \textbf{R2} & \textbf{R3} & \textbf{R1} & \textbf{R2} & \textbf{R3} \\
    \midrule
    \rowcolor{blue}
    & w/o Temporal & 64.9 & 85.9 & 93.8 & 67.6 & 88.1 & 96.6 \\
    \rowcolor{bluegray}
    & w/o Spatial & 62.2 & 84.3 & 94.9 & 61.3 & \textcolor{textblue}{\textbf{89.3}} & 97.8 \\
    \rowcolor{blue}
    \multirow{-3}{*}{Modality Ablation} & w/o Text & 65.9 & 85.7 & \textcolor{textblue}{\textbf{96.4}} & 63.8 & 88.9 & 97.3 \\
    \addlinespace
    \rowcolor{bluegray}
    & w/o Area & 58.9 & 80.3 & 92.9 & 56.3 & 84.7 & 96.3 \\
    \rowcolor{blue}
    & w/o Clarity & 61.4 & 84.9 & 94.3 & 62.1 & 87.9 & 96.6 \\
    \rowcolor{bluegray}
    \multirow{-3}{*}{Spatial Feature} & w/o Centrality & 59.5 & 80.5 & 93.5 & 61.6 & 82.5 & 95.0 \\
    \addlinespace
    \rowcolor{blue}
    & w/o Action & 65.7 & 85.7 & 95.7 & 65.5 & 85.7 & 97.0 \\
    \rowcolor{bluegray}
    & w/o Lip & 55.9 & 77.3 & 89.5 & 57.1 & 80.6 & 95.3 \\
    \rowcolor{blue}
    \multirow{-3}{*}{Temporal Feature} & w/ RAFT & 63.1 & 84.9 & 94.9 & 64.5 & 86.3 & 95.7 \\
    \addlinespace
    \rowcolor{bluegray}
    & w/o Transformer & 65.7 & \textcolor{textblue}{\textbf{87.6}} & 94.9 & 63.9 & 88.5 & 96.2 \\
    \rowcolor{blue}
    & w/o MLP & 66.0 & 84.1 & 94.1 & 63.7 & 82.7 & 94.1 \\
    \rowcolor{bluegray}
    \multirow{-3}{*}{Fusion Strategy} & w/o Gated & 66.2 & 84.1 & 94.1 & 63.4 & 86.3 & 94.5 \\
    \addlinespace
    \rowcolor{blue}
    Learning Strategy & w/o Contrastive & 58.7 & 80.0 & 91.1 & 52.6 & 76.9 & 90.8 \\
    \midrule
    \rowcolor{lastrowcolor}
    \textbf{Complete Model} & \textbf{VIP-Net (Ours)} & \textcolor{textblue}{\textbf{67.3}} & 83.2 & 94.1 & \textcolor{textblue}{\textbf{68.9}} & 84.2 & \textcolor{textblue}{\textbf{97.8}} \\
    \bottomrule
  \end{tabularx}
  \label{tab:ablation_studies}
\end{table}
We establish the superiority of VIP-Net over existing methodologies and proceed to systematically analyze the contributions of individual components to elucidate the underlying sources of this performance advantage.

\textbf{Modality Ablation Study.} Removing the spatial modality leads to a substantial performance drop (62.2\% vs. 64.9\%/65.9\%), with multiple metrics notably lower, indicating its important contribution. In contrast, the text modality exhibits milder degradation, maintaining competitive scores in select metrics. These results suggest that spatial cues play a key role in our model.

	extbf{Feature Component Analysis.} Table~\ref{tab:ablation_studies} reveals the relative importance of feature components: Lip Movement (11.4 point drop when removed) $>$ Area (8.4 points) $>$ Centrality (7.8 points) $>$ Clarity (5.9 points) $>$ Action (1.6 points). This validates our core hypothesis that spatio-temporal patterns jointly determine social importance, with conversational behavior acting as the most critical social cue in human interactions. To verify that lip and action features encode semantic information beyond generic motion, we replace both with optical flow using RAFT~\cite{teed2020raft}. The RAFT-based variant achieves 63.1\% Rank-1, a 4.2-point drop from the full model (67.3\%), confirming that VIP identification requires high-level semantic behavioral understanding rather than low-level raw motion cues.

\textbf{Fusion Strategy Comparison.} Different fusion strategies show varying effectiveness: Gated Fusion (66.2\%) slightly outperforms MLP Fusion (66.0\%), while removing Transformer Fusion reduces performance to 65.7\%. The complete VIP-Net with Transformer Fusion achieves 67.3\%. This demonstrates that while simple concatenation or gating provides reasonable performance, the multi-head attention mechanism in the Transformer is essential for optimally aligning asynchronous spatial and temporal features across the video sequence.

\textbf{Learning Strategy Analysis.} Contrastive Learning shows significant importance, with its removal causing an 8.6-point performance drop (second only to Lip Movement's impact). Mechanistically, the InfoNCE loss explicitly pushes apart the feature embeddings of the true VIP and the ``second most important'' person (the distractor). Without it, the model struggles to establish a clear margin of semantic dominance in crowded scenes, leading to ambiguous rankings.

\textbf{Scene-Specific Performance.} Indoor scenes favor spatial over temporal features: Spatial features improve (67.6\% vs. 64.9\%, 2.7 points higher) while Temporal features decline (61.3\% vs. 62.2\%, 0.9 points lower). This indicates that controlled environments (like offices) enhance appearance-based importance assessment, while diverse, unconstrained scenarios rely more heavily on behavioral dynamics.

\subsection{Hyperparameter Sensitivity}

\begin{figure*}[!t]
  \centering
  \subfloat[Text similarity weight $\lambda_{\text{text}}$]{\includegraphics[width=0.32\textwidth]{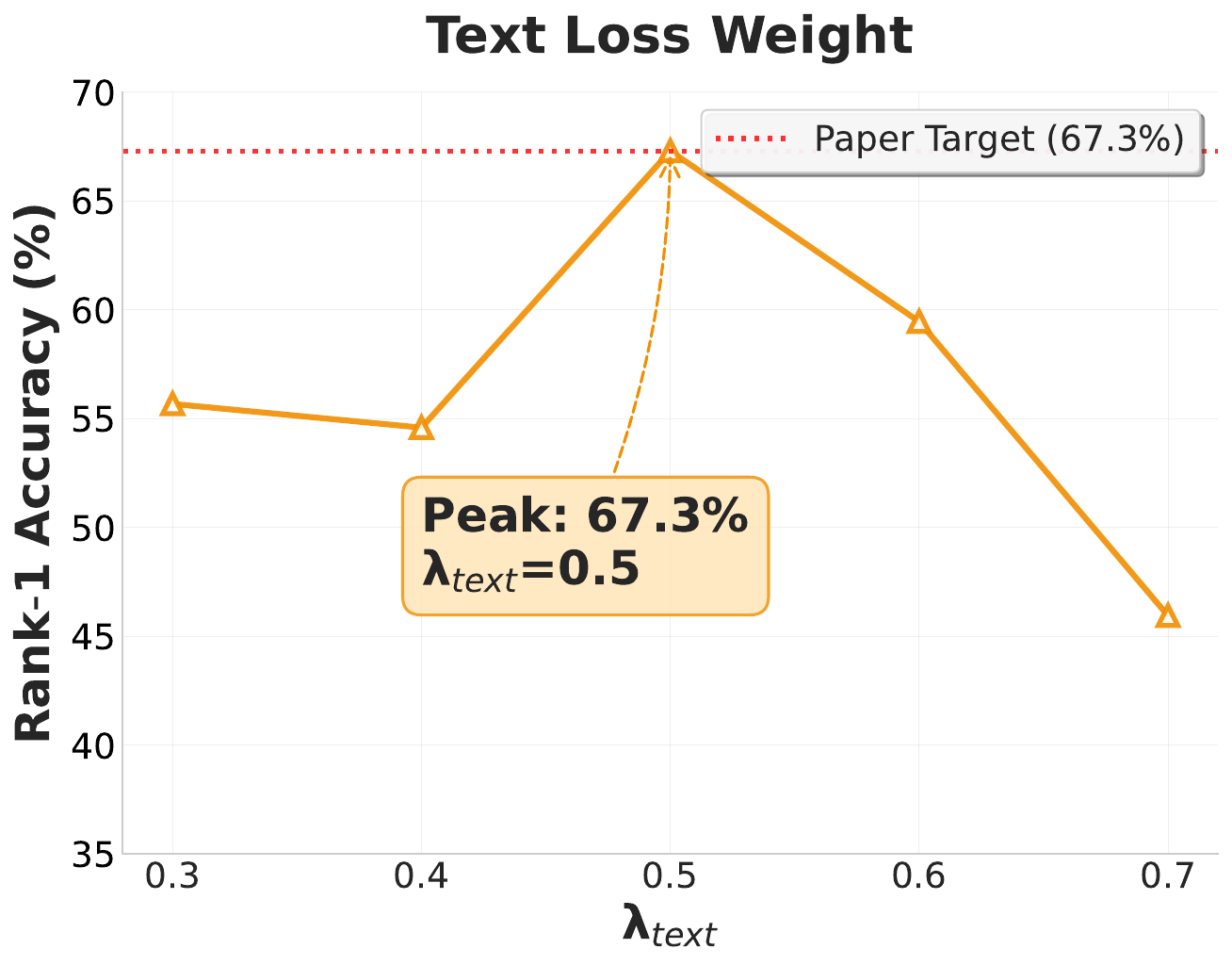}}
  \hfill
  \subfloat[Contrastive weight $\lambda_{\text{cont}}$]{\includegraphics[width=0.32\textwidth]{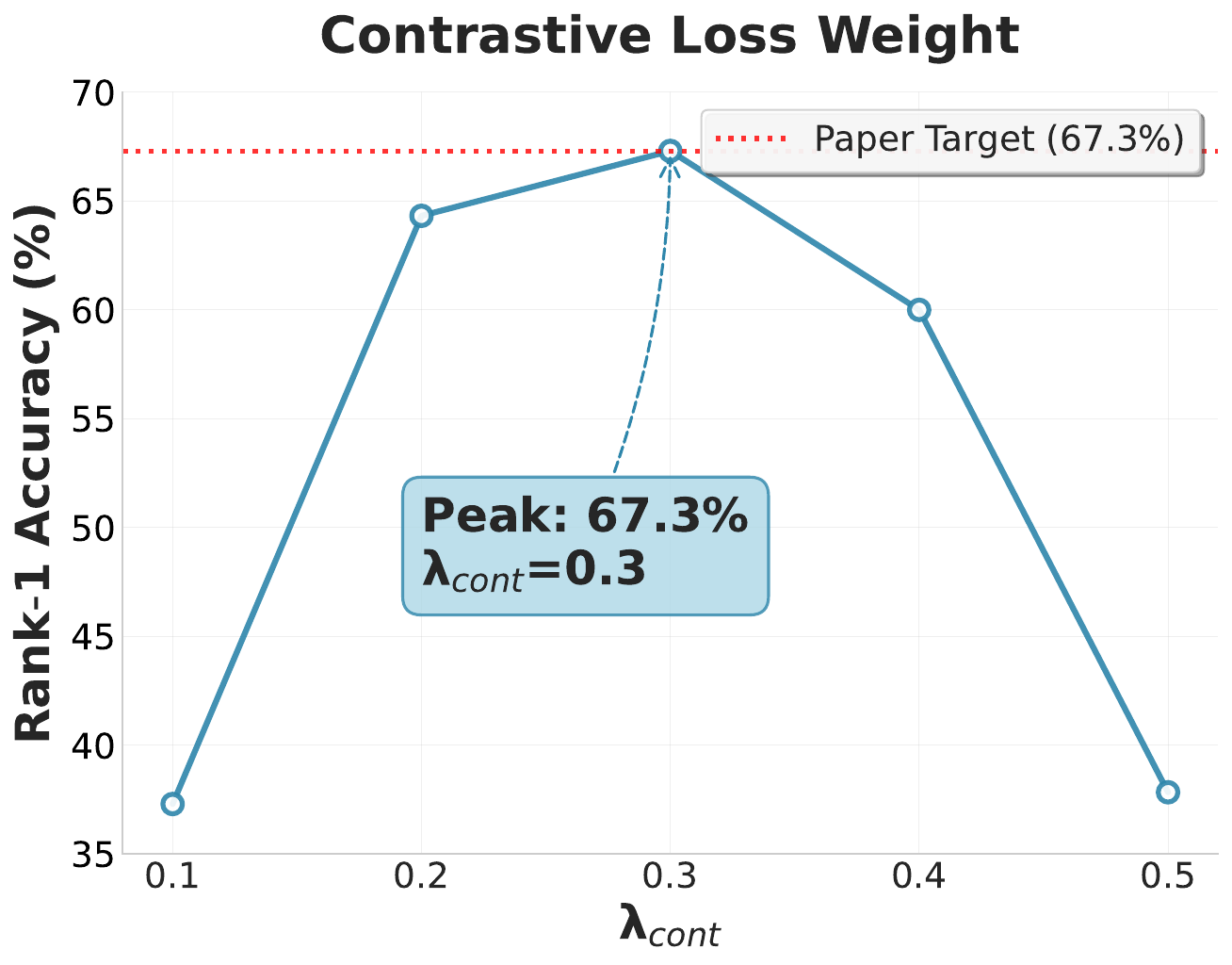}}
  \hfill
  \subfloat[Regularization weight $\lambda_{\text{reg}}$]{\includegraphics[width=0.32\textwidth]{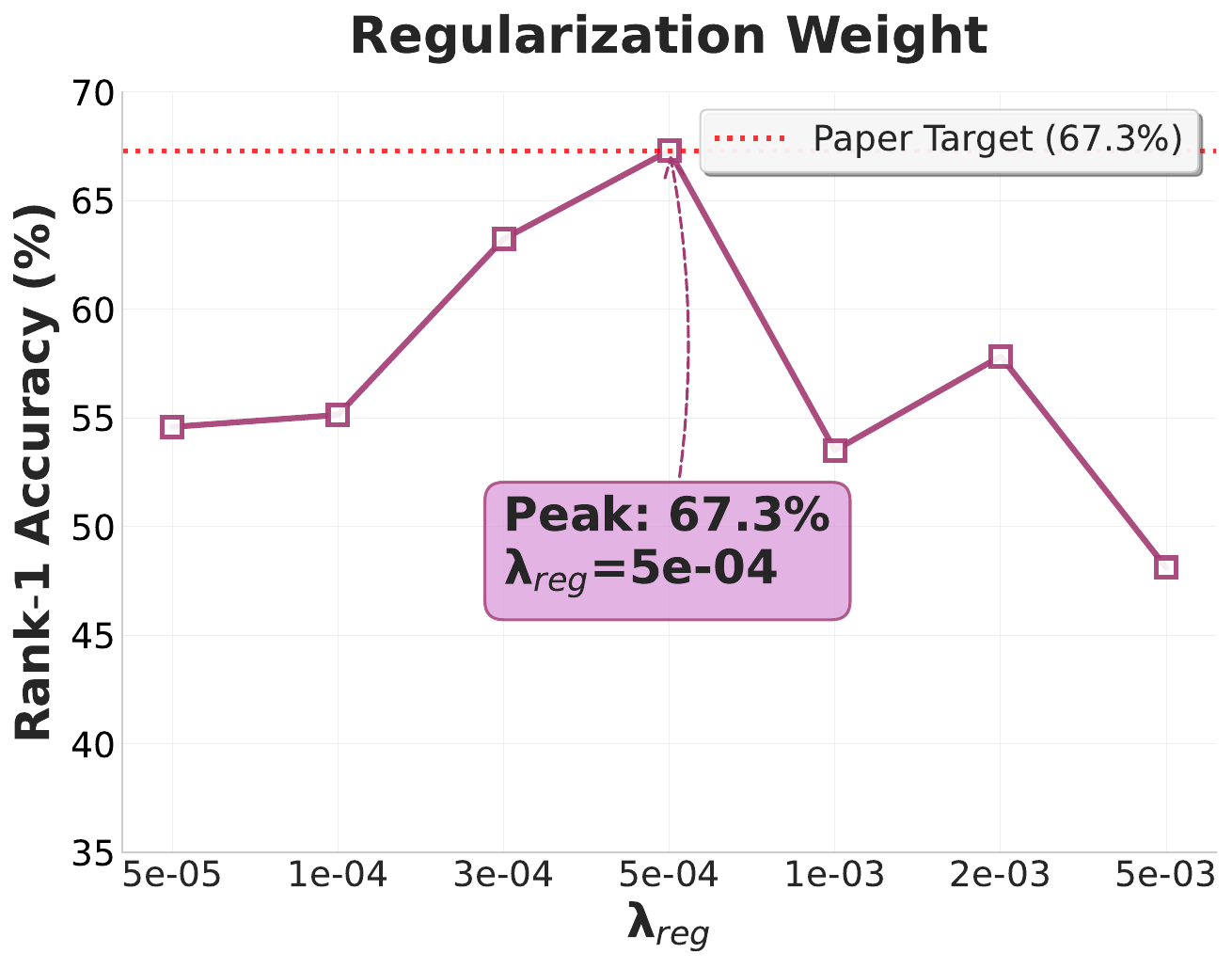}}
  \caption{\textbf{Hyperparameter sensitivity analysis.} Rank-1 accuracy (\%) across different weight values for (a) text similarity weight $\lambda_{\text{text}}$, (b) contrastive learning weight $\lambda_{\text{cont}}$, and (c) regularization weight $\lambda_{\text{reg}}$. Peak performance is marked for each parameter.}
  \label{fig:hyperparameter_analysis}
\end{figure*}

We conduct systematic analysis to evaluate the sensitivity of the three learnable weights ($\lambda_{\mathrm{text}}$, $\lambda_{\mathrm{cont}}$, $\lambda_{\mathrm{reg}}$) in our multi-component loss function. Each weight is varied while keeping others at optimal values, measuring Rank-1 accuracy on the validation split. Fig.~\ref{fig:hyperparameter_analysis} illustrates the results.

\textbf{Text Similarity Weight ($\lambda_{\text{text}}$):} Optimal at 0.5, with moderate sensitivity. Performance peaks at 0.5 and degrades gracefully to 62--65\% Rank-1 accuracy over the range 0.1--0.9. Insufficient weighting ($\lambda_{\text{text}} < 0.3$) reduces description quality, while excessive weighting yields diminishing returns.

\textbf{Contrastive Learning Weight ($\lambda_{\text{cont}}$):} The most critical parameter, optimal at 0.3. High sensitivity causes performance to drop to $\sim$37\% at extremes (0.1 or 0.5), indicating the need for precise balancing to avoid feature space collapse or overpowering the classification loss.

\textbf{Regularization Weight ($\lambda_{\text{reg}}$):} Optimal at 0.0005, with a narrow effective range. Performance degrades above 0.001 due to over-regularization, highlighting the trade-off between overfitting prevention and model expressiveness.

The optimal configuration ($\lambda_{\text{text}}=0.5$, $\lambda_{\mathrm{cont}}=0.3$, $\lambda_{\mathrm{reg}}=0.0005$) achieves 67.3\% Rank-1 accuracy. Notably, contrastive learning requires careful tuning, while text similarity and regularization are more forgiving.

\subsection{Description Quality Evaluation}
Beyond mere localization, a significant contribution of our VIP-Net is its capacity to articulate the underlying reasons for identifying an individual as important. The generation of coherent and contextually grounded rationales fosters transparency and cultivates user trust in the decision-making processes of the model. We assess the linguistic quality and semantic accuracy of these explanations in relation to established ground-truth rationales, as presented in Table~\ref{tab:explanation_quality}.

To rigorously quantify the rationale quality, we employ the Sentence-BERT (SBERT)~\cite{reimers2019sentence} cosine similarity metric. This metric maps both the generated text and the ground truth into a dense semantic space, effectively capturing the nuanced meaning of social narratives beyond simple lexical overlap.

We compare three rationale generation paradigms:
\begin{enumerate}[leftmargin=*]
    \item \textbf{Baseline (Rule-based Templates):} Utilizes predefined linguistic templates populated directly with the discrete feature cues (e.g., spatial prominence, speech activity) extracted by VIP-Net. While highly interpretable and grounded, these rigid templates score the lowest (0.3894 Mean Sim.) due to their inability to produce fluid conversational phrasing or synthesize complex narrative contexts.
    \item \textbf{Unguided LLM:} Inputs only the raw video tokens and the predicted VIP bounding box trajectory into an MLLM. It improves the semantic similarity to 0.6020 by leveraging the vast pre-trained linguistic expressiveness of the model. However, without explicit structural guidance, the MLLM struggles to align its open-ended generation with the precise physical cues that triggered the initial VIP prediction, occasionally leading to narrative hallucination.
    \item \textbf{Guided LLM (Ours):} Our proposed feature-guided refinement module integrates the raw video, the predicted VIP trajectory, \textit{and} the structured multi-dimensional feature cues extracted by VIP-Net. This hybrid mechanism acts as a robust semantic anchor. Consequently, it achieves the highest semantic alignment (0.6333 Mean Sim., a 62.7\% relative improvement over the baseline). This demonstrates that feeding deterministic feature rankings into an LLM successfully bridges the gap between low-level visual perception and high-level cognitive explanation.
\end{enumerate}

\begin{table}[!t]
 \caption{Comparison of description quality across different variants. Baseline uses structured descriptions generated from predicted VIP ID and ranked feature cues. Unguided LLM receives predicted VIP ID and raw video. Guided LLM additionally incorporates ranked feature cues. SBERT similarity measures semantic alignment, variance indicates consistency. Guided LLM achieves the highest mean similarity and improves all samples, outperforming unguided LLM and baseline.}
    \centering
    \footnotesize
    \setlength{\tabcolsep}{1pt}
  \begin{tabularx}{\linewidth}{l c *{3}{>{\centering\arraybackslash}X}}
        \toprule
        \rowcolor{headercolor}
        \multirow{1}{*}{\textbf{Variant}} & \multirow{1}{*}{\textbf{Inputs}} & \multirow{1}{*}{\textbf{Mean Sim.}} & \textbf{Variance Sim.} & \textbf{$\Delta$ vs. Baseline} \\
        \midrule
        \rowcolor{purple}
    Baseline & id,cues & 0.3894 & 0.1008 & \textemdash{} \\
        \rowcolor{purplegray}
        \multirow{2}{*}{unguided LLM} & \multirow{2}{*}{id,video} & \multirow{2}{*}{0.6020} & \textcolor{textblue}{\multirow{2}{*}{\textbf{0.0747}}} & $\uparrow\,0.2126$ (+54.6\%) \\
        \midrule
        \rowcolor{lastrowcolor}
        \multirow{2}{*}{\textbf{guided LLM}} & \multirow{2}{*}{id,video,cues} & \textcolor{textblue}{\multirow{2}{*}{\textbf{0.6333}}} & \multirow{2}{*}{0.0852} & \textcolor{textblue}{\textbf{$\uparrow\,0.2440$ (+62.7\%)}} \\
        \bottomrule
    \end{tabularx}
 \label{tab:explanation_quality}
\end{table}

\begin{figure}[!t]
  \centering
  \subfloat[Case 1: In the classroom, the model identifies the standing student as the key individual, visibly contrasting with seated peers.\label{fig:case1}]{\includegraphics[width=\linewidth]{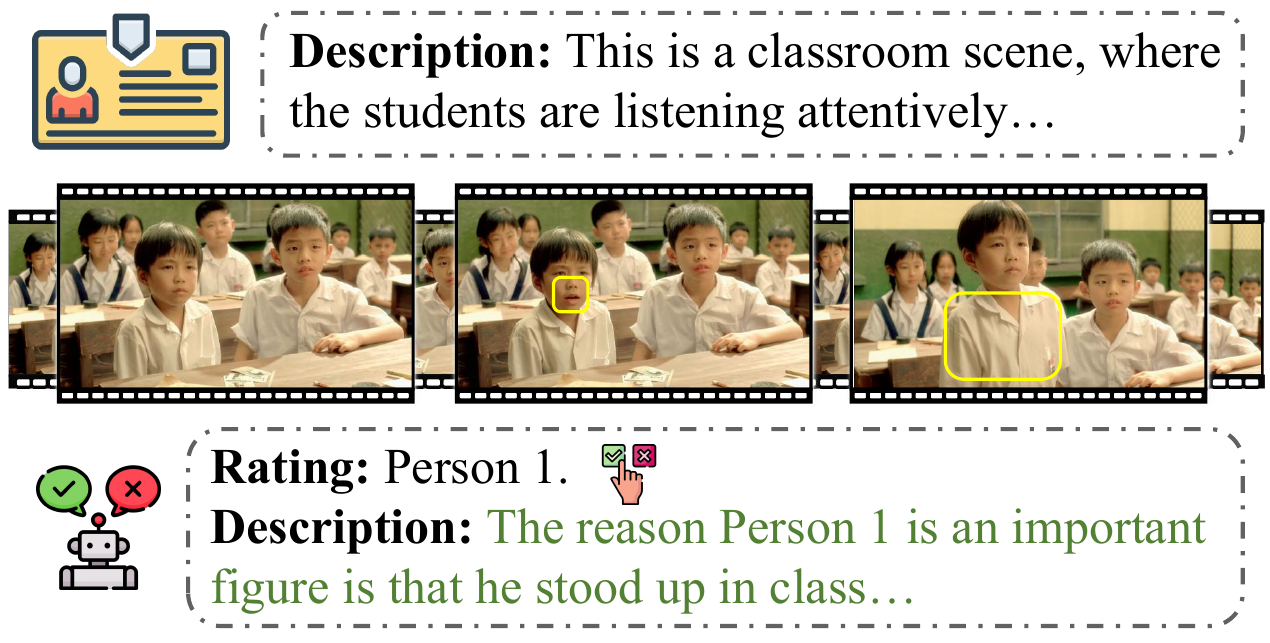}}
  \hfil
  \subfloat[Case 2: In the courtroom, the model detects the principal figure in the corner, whose action draws collective attention.\label{fig:case2}]{\includegraphics[width=\linewidth]{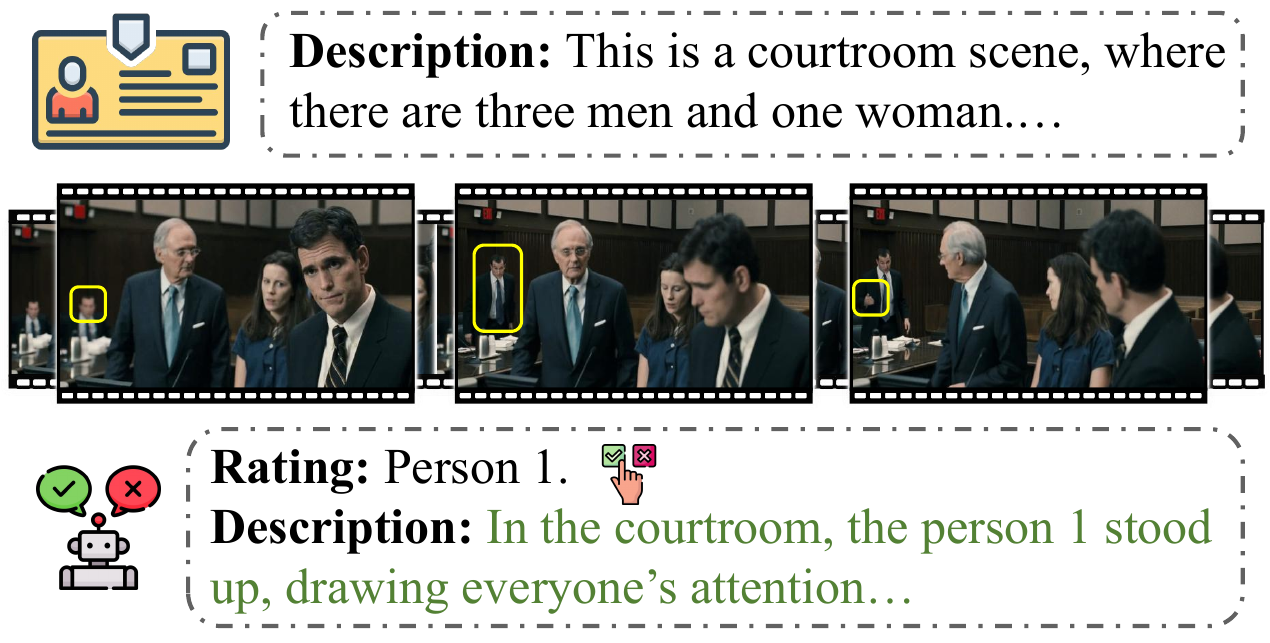}}
  \caption{\textbf{Qualitative examples}. This topic not only identifies key individuals but also elucidates the rationale behind their significance. Such elaboration enhances the understanding of the scene by providing contextual information that supports the interpretation of the visual content.}
  \label{fig:explanation_case}
\end{figure}

We evaluate three variants of description generation by benchmarking their performance using SBERT cosine similarity on the 1,850 validation videos where VIP-Net predicts the correct individual. A single checkpoint with the best validation Rank-1 score is frozen for evaluation. We first reconstruct the structured description from ranked feature cues and then optionally refine it with a local LLM that receives the video clip together with this initial description. The three settings correspond to using the structured description alone, the LLM without additional cues, and the LLM supplied with our feature hints. As shown in Table~\ref{tab:explanation_quality}, the unguided LLM improves the average SBERT similarity over the structured description baseline and reduces variance, indicating that the second-stage model benefits from direct visual evidence. Incorporating our ranked cues yields a further 0.03 improvement over the unguided LLM and enhances all samples in the guided setting, achieving an overall similarity of 0.63 with a variance of 0.09. Fig.~\ref{fig:explanation_case} shows representative examples where the model correctly identifies the VIP and provides descriptions.

\section{Conclusion}
\label{sec:conclusion}
In this paper, we formally extend the frontier of important person identification from static images to dynamic video environments. We introduce and formulate the inherent challenge of Temporal Importance Shift (TIS)---demonstrating that an individual's social prominence is rarely static and often evolves through complex interactions over time. To support rigorous research in this domain, we present \textbf{Temporal-VIP}, to the best of our knowledge, the first large-scale, fine-grained benchmark equipped with multi-dimensional rationale annotations.
To address the limitations of existing spatial and pure tracking-based paradigms, we propose \textbf{VIP-Net}, a specialized multi-modal framework. VIP-Net distinctly decouples the task into relational spatio-temporal dynamics modeling and feature-guided semantic rationale generation. Extensive empirical evaluations not only demonstrate VIP-Net's substantial superiority over state-of-the-art multi-modal variants (achieving 67.3\% Rank-1 accuracy, a significant 13.4\% absolute gain), but also yield profound diagnostic insights. Crucially, our experiments prove that while Large Vision-Language Models possess vast semantic knowledge, they require deterministic, physical interaction cues as structural anchors to prevent reasoning hallucinations. 
Ultimately, this study bridges the cognitive gap between low-level temporal tracking and high-level social scene understanding. Future work will naturally extend this paradigm to unconstrained, highly crowded environments (e.g., massive surveillance networks) and explore cross-camera cooperative social inference, further pushing the boundaries of explainable in human-centric analysis.

% \section*{Acknowledgments}
% Journal style: unnumbered.
% (Keep empty for anonymous review, or fill in for camera-ready.)

% -------------------------
% Supplementary material
% NOTE: Per request, supplementary is kept as a separate, standalone file.
% The following appendix inclusion is intentionally disabled for now.
% -------------------------
% \appendices
% \section{Supplementary Material}
% \label{sec:supplementary}
% \input{supplementary_content}

\bibliographystyle{IEEEtran}
\bibliography{main}

\end{document}